\documentclass[runningheads]{llncs}
\usepackage{graphicx}
\usepackage{amsmath,amssymb} % define this before the line numbering.
\usepackage{color}
\newcommand{\ve}[1]{\mbox{{\bf #1}}} % for displaying a vector or matrix
\usepackage{cite}

\usepackage{array}

\definecolor{myGreen}{rgb}{0, .8, .3}
\definecolor{myRed}{rgb}{0.8, .2, .2}

\definecolor{renjiao}{RGB}{0,139,139}

\usepackage[width=122mm,left=12mm,paperwidth=146mm,height=193mm,top=12mm,paperheight=217mm]{geometry}
\begin{document}
% \linenumbers
\pagestyle{headings}
\mainmatter
\def\ECCV18SubNumber{409}  % Insert your submission number here

\title{Faces as Lighting Probes via Unsupervised Deep Highlight Extraction} % Replace with your title

\titlerunning{Faces as Lighting Probes via Unsupervised Deep Highlight Extraction}

\authorrunning{R. Yi et al. }

\author{Renjiao Yi\inst{1,2}, Chenyang Zhu\inst{1,2}, Ping Tan\inst{1}, Stephen Lin\inst{3}}

%Please write out author names in full in the paper, i.e. full given and family names. 
%If any authors have names that can be parsed into FirstName LastName in multiple ways, please include the correct parsing, in a comment to the volume editors:
%\index{Lastnames, Firstnames}
%(Do not uncomment it, because you may introduce extra index items if you do that...)

\institute{
	%Department,\\
	Simon Fraser University, Burnaby, Canada\\
	\email{\{renjiaoy, cza68, pingtan\}@sfu.ca}\and
	National University of Defense Technology, Changsha, China\and
	Microsoft Research, Beijing, China\\
	%\email{ \{author1,author2\}@univ.edu}
	\email{stevelin@microsoft.com}
}

\maketitle

\begin{abstract}
We present a method for estimating detailed scene illumination using human faces in a single image. In contrast to previous works that estimate lighting in terms of low-order basis functions or distant point lights, our technique estimates illumination at a higher precision in the form of a non-parametric environment map. Based on the observation that faces can exhibit strong highlight reflections from a broad range of lighting directions, we propose a deep neural network for extracting highlights from faces, and then trace these reflections back to the scene to acquire the environment map. Since real training data for highlight extraction is very limited, we introduce an unsupervised scheme for finetuning the network on real images, based on the consistent diffuse chromaticity of a given face seen in multiple real images. In tracing the estimated highlights to the environment, we reduce the blurring effect of skin reflectance on reflected light through a deconvolution determined by prior knowledge on face material properties. Comparisons to previous techniques for highlight extraction and illumination estimation show the state-of-the-art performance of this approach on a variety of indoor and outdoor scenes.
\end{abstract}

\section{Introduction}
Spicing up selfies by inserting virtual hats, sunglasses or toys has become easy to do with mobile augmented reality (AR) apps like {\it Snapchat} \cite{Snapchat}. But while the entertainment value of mobile AR is evident, it is just as clear to see that the generated results are usually far from realistic. A major reason is that virtual objects are typically not rendered under the same illumination conditions as in the imaged scene, which leads to inconsistency in appearance between the object and its background. For high photorealism in AR, it is thus necessary to estimate the illumination in the image, and then use this estimate to render the inserted object compatibly with its surroundings.
%\begin{figure}
%\centering
%\includegraphics[width=1 \linewidth]{Figures/teaser-1.PNG}\
%   \caption{{\color{renjiao}{We present an approach to use CNN to extract highlight layer from input LDR face images, then recover the high frequency illumination from extracted highlight layers. The recovered illumination can be used to create a more realistic object insertion then previous methods using low-order Spherical Harmonics illumination without ability to separate shading color from intrinsic face color. The illumination color can be successfully recovered by our proposed pipeline. }}
%}
%\label{fig:teaser}
%\end{figure}

Illumination estimation from a single image is a challenging problem because lighting is intertwined with geometry and reflectance in the appearance of a scene. To make this problem more manageable, most methods assume the geometry and/or reflectance to be known \cite{sato1999acquiring, ramamoorthi2001signal, wang2002estimation, sato2003illumination, li2003multiple, pessoa2010photorealistic, panag2011illumination, lombardi2016reflectance}. Such knowledge is generally unavailable in practice; however, there exist priors about the geometry and reflectance properties of human faces that have been exploited for illumination estimation \cite{kemel2011face, knorr2014real, li2014intrinsic, richardson2017learning}. Faces are a common occurrence in photographs and are the focus of many mobile AR applications. The previous works on face-based illumination estimation consider reflections to be diffuse and estimate only the low-frequency component of the environment lighting, as diffuse reflectance acts as a low-pass filter on the reflected illumination \cite{ramamoorthi2001signal}. However, a low-frequency lighting estimate often does not provide the level of detail needed to accurately depict virtual objects, especially those with shiny surfaces. %as shown in Figure~\ref{fig:pipeline}.

In addressing this problem, we consider the parallels between human faces and mirrored spheres, which are conventionally used as lighting probes for acquiring ground truth illumination. What makes a mirrored sphere ideal for illumination recovery is its perfectly sharp specular reflections over a full range of known surface normals. Rays can be traced from the camera's sensor to the sphere and then to the surrounding environment to obtain a complete environment map that includes lighting from all directions and over all frequencies, subject to camera resolution. We observe that faces share these favorable properties to a large degree. They produce fairly sharp specular reflections (highlights) over its surface because of the oil content in skin. Moreover, faces cover a broad range of surface normals, and there exist various methods for recovering face geometry from a single image \cite{blanz1999morphable, kemel2011face, yang2011expression, richardson2017learning, sela2017unrestricted}. Unlike mirrored spheres, the specular reflections of faces are not perfectly sharp and are mixed with diffuse reflection. In this paper, we propose a method for dealing with these differences to facilitate the use of faces as lighting probes.

We first present a deep neural network for separating specular highlights from diffuse reflections in face images. The main challenge in this task is the lack of ground truth separation data on real face images for use in network training. Although ground truth separations can be generated synthetically using graphics models \cite{shi2017learning}, it has become known that the mismatch between real and synthetic data can lead to significant reductions in performance \cite{shrivastava2017learning}. We deal with this issue by pretraining our network with synthetic images and then finetuning the network using an unsupervised strategy with real photos. Since there is little real image data on ground truth separations, we instead take advantage of the property that the diffuse chromaticity values over a given person's face are relatively unchanged from image to image, aside from a global color rescaling due to different illumination colors and sensor attributes. From this property, we show that the diffuse chromaticity of multiple aligned images of the same face should form a low-rank matrix. We utilize this low-rank feature in place of ground truth separations to finetune the network using multiple real images of the same face, downloaded from the MS-celeb-1M database \cite{guo2016ms}. This unsupervised finetuning is shown to significantly improve highlight separation over the use of supervised learning on synthetic images alone.

With the extracted specular highlights, we then recover the environment illumination. This recovery is inspired by the frequency domain analysis of reflectance in \cite{ramamoorthi2001signal}, which concludes that reflected light is a convolved version of the environment map. Thus, we estimate illumination through a deconvolution of the specular reflection, in which the deconvolution kernel is determined from prior knowledge of face material properties. This approach enables recovery of higher-frequency details in the environment lighting. 

This method is validated through experimental comparisons to previous techniques for highlight extraction and illumination estimation. On highlight extraction, our method is shown to produce results that more closely match the ground truth acquired by cross-polarization. For illumination estimation, greater precision is obtained over a variety of both indoor and outdoor scenes. We additionally show that the 3D positions of local point lights can be estimated using this method, by triangulating the light source positions from the environment maps of multiple faces in an image. With this 3D lighting information, the spatially variant illumination throughout a scene can be obtained. Recovering the detailed illumination in a scene not only benefits AR applications but also can promote scene understanding in general.

\begin{figure*}[t] 
\centering
\includegraphics[width=0.93 \linewidth]{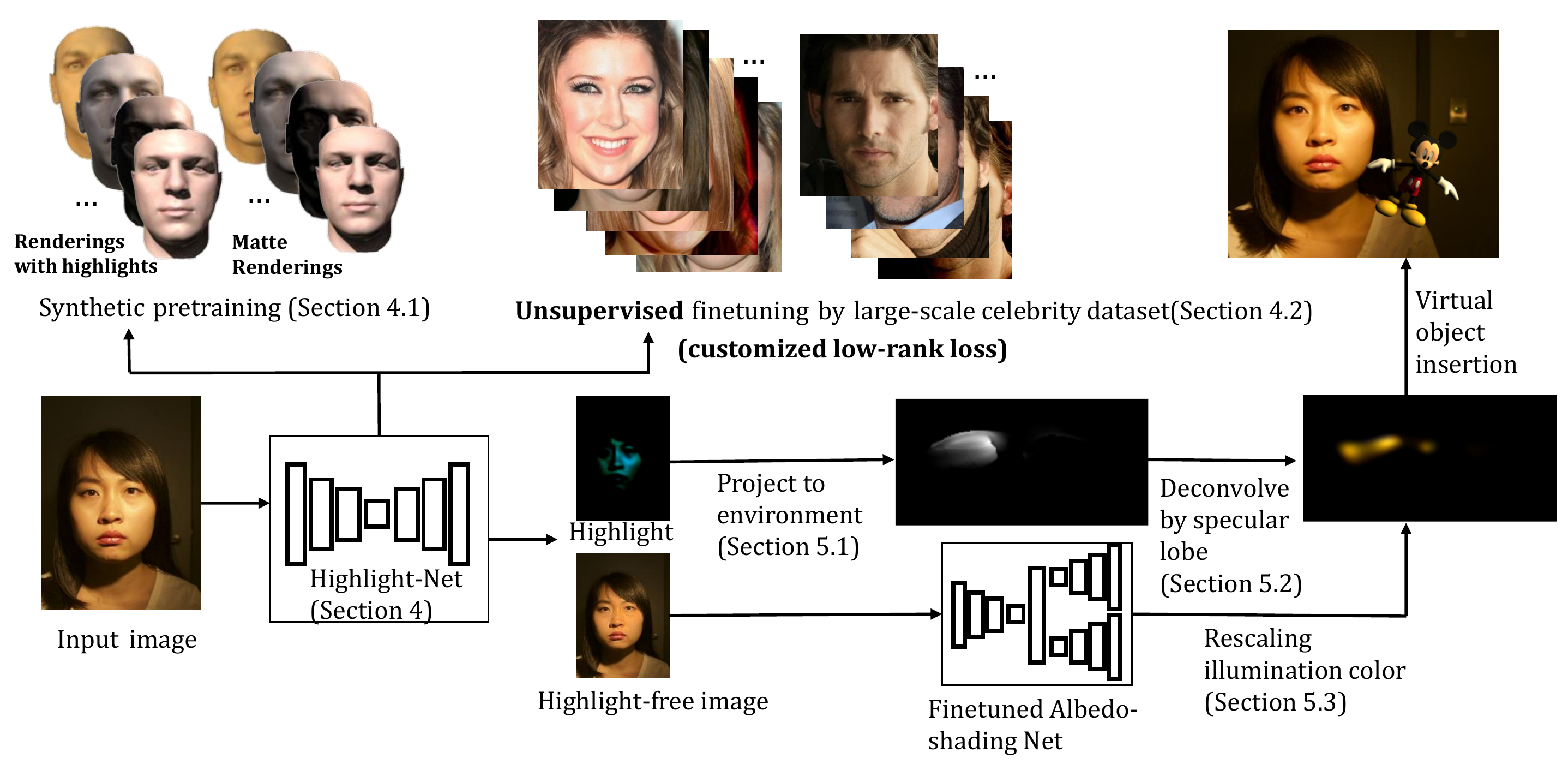}\
   \caption{Overview of our method. An input image is first separated into its highlight and diffuse layers. We trace the highlight reflections back to the scene according to facial geometry to recover a non-parametric environment map. A diffuse layer obtained through intrinsic component separation \cite{narihira2015direct} is used to determine illumination color. With the estimated environment map, virtual objects can be inserted into the input image with consistent lighting. 
}
\label{fig:pipeline}
\end{figure*}
\section{Related work}
  
\noindent\textbf{Highlight extraction}
involves separating the diffuse and specular reflection components in an image. This problem is most commonly addressed by removing highlights with the help of chromatic \cite{tan2004separating, tan2005reflection, yang2010real, kim2013specular} as well as spatial \cite{tan2003highlight, tan2006separation, mallick2006specularity} information from neighboring image areas, and then subtracting the resulting diffuse image from the original input to obtain the highlight component. These techniques are limited in the types of surface textures that can be handled, and they assume that the illumination color is uniform or known.

In recent work \cite{li2017specular}, these restrictions are avoided for the case of human faces by utilizing additional constraints derived from physical and statistical face priors. Our work also focuses on human faces but employs a deep learning approach instead of a physics-based solution for highlight extraction. While methods developed from physical models have a tangible basis, they might not account for all factors that influence image appearance, and analytical models often provide only a simplified approximation of natural mechanisms. In this work, we show that directly learning from real image data can lead to improved results that additionally surpass deep learning on synthetic training data \cite{shi2017learning}.

\noindent\textbf{Illumination estimation}
is often performed from a single image, as this is the only input available in many applications. The majority of single-image methods assume known geometry in the scene and estimate illumination from shading \cite{ramamoorthi2001signal, wang2002estimation, li2003multiple, pessoa2010photorealistic} and shadows \cite{sato1999acquiring, sato2003illumination, li2003multiple, okabe2004spherical, panag2011illumination}. Some methods do not require geometry to be known in advance, but instead they infer this information from the image by employing priors on object geometry \cite{barron2015shape,lopez2010compositing,nishino2004eyes} or by fitting shape models for faces \cite{kemel2011face, garrido2013reconstructing, knorr2014real, li2014intrinsic, richardson2017learning}. Our work also makes use of statistical face models to obtain geometric information for illumination estimation.

An illumination environment can be arbitrarily complex, and nearly all previous works employ a simplified parametric representation as a practical approximation. Earlier techniques mainly estimate lighting as a small set of distant point light sources \cite{sato1999acquiring, wang2002estimation, sato2003illumination, li2003multiple, panag2011illumination}. More recently, denser representations in the form of low-order spherical harmonics \cite{ramamoorthi2001signal, kemel2011face, garrido2013reconstructing, knorr2014real, li2014intrinsic, barron2015shape, richardson2017learning} and Haar wavelets \cite{okabe2004spherical} have been recovered. The relatively small number of parameters in these models simplifies optimization but provides limited precision in the estimated lighting. A more detailed lighting representation may nevertheless be infeasible to recover from shading and shadows because of the lowpass filtering effect of diffuse reflectance \cite{ramamoorthi2001signal} and the decreased visibility of shadow variations under extended lighting.

Greater precision has been obtained by utilizing lighting models specific to a certain type of scene. For outdoor environments, sky and sun models have been used for accurate recovery of illumination \cite{lalonde2008does, lalonde2010sun, hold2016deep,calian2018faces}. %\renjiao{For indoor scenes, classifiers are trained to recognize in-view light sources in \cite{karsch2014automatic}.} 
In research concurrent to ours, indoor illumination is predicted using a convolutional neural network trained on data from indoor environment maps \cite{gardner2017learning}. Similar to our work, it estimates a non-parametric representation of the lighting environment with the help of deep learning. Our approach differs in that it uses human faces to determine the environment map, and employs deep learning to recover an intermediate quantity, namely highlight reflections, from which the lighting can be analytically solved. Though our method has the added requirement of having a face in the image, it is not limited to indoor scenes and it takes advantage of more direct evidence about the lighting environment. We later show that this more direct evidence can lead to higher precision in environment map estimates.

Highlight reflections have been used together with diffuse shading to jointly estimate non-parametric lighting and an object's reflectance distribution function \cite{lombardi2016reflectance}. In that work, priors on real-world reflectance and illumination are utilized as constraints to improve inference in an optimization-based approach. The method employs an object with known geometry, uniform color, and a shiny surface as a probe for the illumination. By contrast, our work uses arbitrary faces, which are a common occurrence in natural scenes. As shown later, the optimization-based approach can be sensitive to the complications presented by faces, such as surface texture, inexact geometry estimation, and spatially-variant reflectance. Our method reliably extracts a key component of illumination estimation -- highlight reflections -- despite these obstacles by using a proposed deep learning scheme.

\section{Overview}

As shown in Figure~\ref{fig:pipeline}, we train a deep neural network called {\it Highlight-Net} to extract the highlight component from a face image. This network is trained in two phases. First, pretraining is performed with synthetic data (Section \ref{sec:pretraining}). Subsequently, the network is finetuned in an unsupervised manner with real images from a celebrity dataset (Section \ref{sec:finetuning}). 

For testing, the network takes an input image and estimates its highlight layer. Together with reconstructed facial geometry, the extracted highlights are used to obtain an initial environment map, by tracing the highlight reflections back towards the scene. This initial map is blurred due to the band-limiting effects of surface reflectance \cite{ramamoorthi2001signal}. To mitigate this blur, our method performs deconvolution on the environment map using kernels determined from facial reflectance statistics (Section \ref{sec:illumination}).

\section{Face highlight removal}\label{sec:removal}

%In this section, we describe the details of training the Highlight-Net, where Section \ref{sec:pretraining} explains the pretraining, and Section \ref{sec:finetuning} details the unsupervised finetuning with a customized low rank loss layer. 

%\begin{figure}
%\centering
%\includegraphics[width=1 \linewidth]{Figures/highlightnet.PNG}\
%   \caption{Highlight-Net structure. }
%\label{fig:highlightnet}
%\end{figure}

\subsection{Pretraining with synthetic data}
\label{sec:pretraining}
For Highlight-Net, we adopt a network structure used previously for intrinsic image decomposition \cite{narihira2015direct}, a related image separation task. 
%\textcolor{red}{as illustrated in Figure \ref{fig:highlightnet} [remove this figure?]}
To pretrain this network, we render synthetic data using generic face models \cite{paysan20093d} and real indoor and outdoor HDR environment maps collected from the Internet. Details on data preparation are presented in Section~\ref{sec:data}. With synthetic ground truth specular images, we minimize the L2 loss between the predicted and ground truth highlights for pretraining. 

%-------------------------------------------------------------------------
\subsection{Unsupervised finetuning on real images}
\label{sec:finetuning}

With only pretraining on synthetic data, Highlight-Net performs inadequately on real images. This may be attributed to the limited variation of face shapes, textures, and environment maps in the synthetic data, as well as the gap in appearance between synthetic and real face images. Since producing a large-scale collection of real ground-truth highlight separation data is impractical, we present an unsupervised strategy for finetuning Highlight-Net that only requires real images of faces under varying illumination environments.

This strategy is based on the observation that the diffuse chromaticity over a given person's face should be consistent in different images, regardless of illumination changes, because a person's facial surface features should remain the same. Among images of the same face, the diffuse chromaticity map should differ only by global scaling factors determined by illumination color and sensor attributes, which we correct in a preprocessing step. Thus, a matrix constructed by stacking the aligned diffuse chromaticity maps of a person should be of low rank. In place of ground-truth highlight layers of real face images, we use this low-rank property of ground-truth diffuse layers to finetune our Highlight-Net. 
%Our method thus aligns multiple images of the same person, separates their diffuse and highlight layers using Highlight-Net, constructs a diffuse chromaticity matrix, and enforces a low-rank constraint to finetune Highlight-Net.

\begin{figure}[t]
\centering
\begin{tabular}
{>{\centering\arraybackslash}m{0.45\linewidth}
>{\centering\arraybackslash}m{0.45\linewidth}}
\multicolumn{2}{c}{\includegraphics[width=0.9\linewidth]{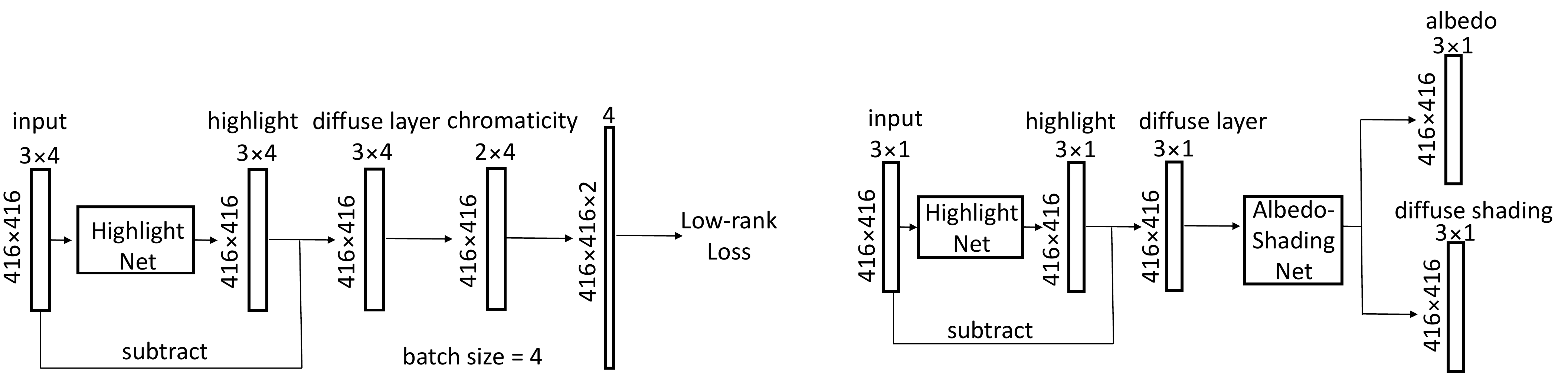}}\\
(a) & (b)\\
\end{tabular}
   \caption{(a) Network structure for finetuning Highlight-Net; (b) Testing network structure for separating an input face image into three layers: highlight, diffuse shading, and albedo.}
\label{fig:network2}
\end{figure}

This finetuning is implemented using the network structure shown in Figure~\ref{fig:network2} (a), where Highlight-Net is augmented with a low-rank loss. The images for training are taken from the MS-celeb-1M database \cite{guo2016ms}, which contains 100 images for each of 100,000 celebrities. After some preprocessing described in Section \ref{sec:data}, we have a set of aligned frontal face images under 
%various illuminations 
a consistent illumination color for each celebrity. 

%Some examples of those preprocessed data are showed in Figure~\ref{fig:realface}.  

From this dataset, four face images of the same celebrity are randomly selected for each batch. A batch is fed into Highlight-Net to produce the estimated highlight layers for the four images. These highlight layers are subtracted from the original images to obtain the corresponding diffuse layers. For a diffuse layer $I_d$, its diffuse chromaticity map is computed per-pixel as 
\begin{equation}
\begin{split}
chrom(I_d) = \frac{1}{(I_d(r)+I_d(g)+I_d(b))}\left(I_d(r), I_d(g)\right)\label{equation:chromaticity}
\end{split}
\end{equation}
\noindent where $r$, $g$, and $b$ denote the color channels. Each diffuse chromaticity map is then reshaped into a vector $I^{dc}$, and the vectors of the four images are stacked into a matrix $D=\begin{bmatrix}I_1^{dc},I_2^{dc},I_3^{dc},I_4^{dc}\end{bmatrix}^T$. With a low-rank loss enforced on $D$, Highlight-Net is finetuned through backpropagation.

Since the diffuse chromaticity of a face should be consistent among images, the rank of matrix $D$ should ideally be one. So we define the low-rank loss as its second singular value, during backpropagation the partial derivative of $\sigma_2$ with respect to each matrix element is evaluated according to \cite{papadopoulo2000estimating}:
\begin{equation}
\begin{split}
&D=U\Sigma V^T,\quad\quad
\Sigma=diag(\sigma_1,\sigma_2,\sigma_3,\sigma_4), \\
&loss_{lowrank}=\sigma_2,\quad\quad{\frac{\partial \sigma_2}{\partial D_{i,j}}}=U_{i,2}\times V_{j,2}.\label{equation:lowrank}
\end{split}
\end{equation}

%\begin{equation}
%\begin{split}
%&D=U\Sigma V^T, \Sigma=diag(\sigma_1,\sigma_2,\sigma_3,\sigma_4), 
%{\frac{\partial \sigma_2}{\partial D_{i,j}}}=U_{i,2}\times V_{j,2}.\label{equation:backpropagation}
%\end{split}
%\end{equation}
%During backpropagation, the partial derivative of $\sigma_2$ with respect to each matrix element is evaluated according to \cite{papadopoulo2000estimating}: 
%\begin{equation}
%\begin{split}
%\label{equation:backpropagation}
%\end{split}
%\end{equation}
%\ping{I am curious what if we use the nuclear norm. }
%\steve{The matrix rank would actually be two if the illumination colors are different within a batch. Should we express $D$ as eight stacked vectors (4 images x 2 chromaticity values) and say that the ideal rank should be two?}\renjiao{In the experiments, actually I am putting rgb channels for each chromaticity as one vector, not 3 vectors, so when there are 4 photos, we only have 4 rows in chromaticity matrix, not 4*3=12 rows. }

%\cite{papadopoulo2000estimating}
%In this way, we can compute the partial derivative of $\sigma_2$ with respect to each value in chromaticity matrix $D$. 

\section{Illumination estimation} \label{sec:illumination}
%In this section, we recover the illumination represented by an environment map  from the separated highlight layer. 
%, since face region covers a large range of normal directions, and highlight exists all over the face due to face oils and sweats. 

\begin{figure}[t]
\centering
\includegraphics[width=0.6 \linewidth]{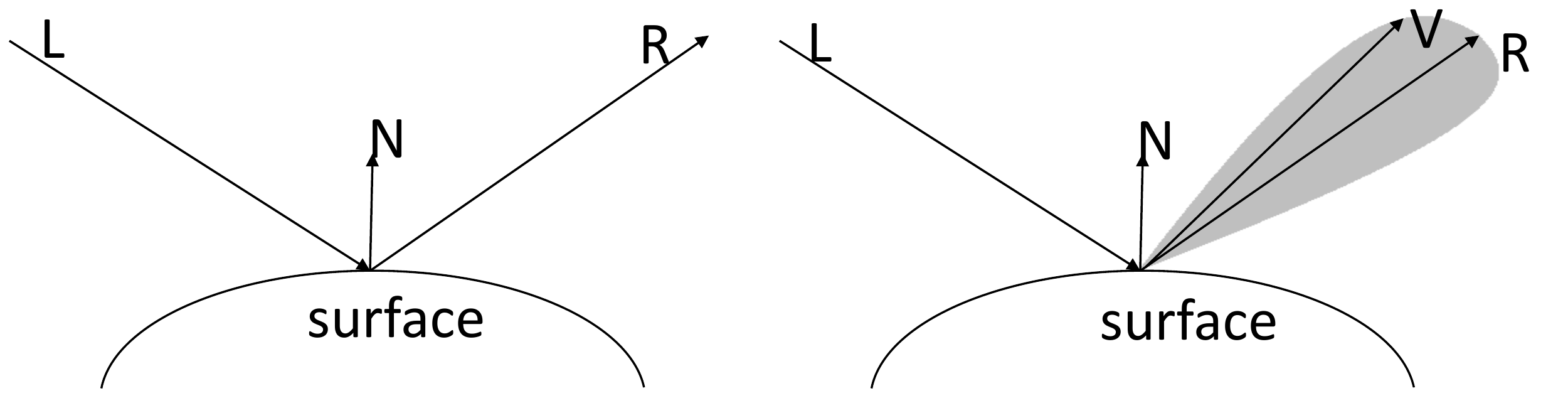}\
   \caption{Left: Mirror reflection. Right: Specular reflection of a rough surface. }
%   Specular reflection concentrates nearby the direction $R$, which is the reflection of incident lighting direction $L$. }
\label{fig:reflections}
\end{figure}

\subsection{Environment map initialization} \label{sec:init}
The specular reflections of a mirror are ideal for illumination estimation, because the observed highlights can be exactly traced back to the environment map when surface normals are known. This exact tracing is possible because a highlight reflection is directed along a single reflection direction $R$ that mirrors the incident lighting direction $L$ about the surface normal $N$, as shown on the left side of Figure~\ref{fig:reflections}. This raytracing approach is widely used to capture environment maps with mirrored spheres in computer graphics applications.

For the specular reflections of a rough surface like human skin, the light energy is instead tightly distributed around the mirror reflection direction, as illustrated on the right side of Figure~\ref{fig:reflections}. This specular lobe can be approximated by the specular term of the Phong model \cite{phong1975illumination} as
 \begin{equation}
 \begin{split}
 I_s = k_s(R \cdot V)^{\alpha },\hspace*{01.0cm}
 R =2(L \cdot N )N -L \label{equation:Phong}
 \end{split}
 \end{equation}
\noindent where $k_s$ denotes the specular albedo, $V$ is the viewing direction, and $\alpha$ represents the surface roughness. We specifically choose to use the Phong model to take advantage of statistics that have been compiled for it, as described later.

\begin{figure}
\centering
\begin{tabular}
{>{\centering\arraybackslash}m{0.25\linewidth}
>{\centering\arraybackslash}m{0.225\linewidth}
>{\centering\arraybackslash}m{0.225\linewidth}
>{\centering\arraybackslash}m{0.25\linewidth}}
\multicolumn{4}{c}
{\includegraphics[width=0.93 \linewidth]{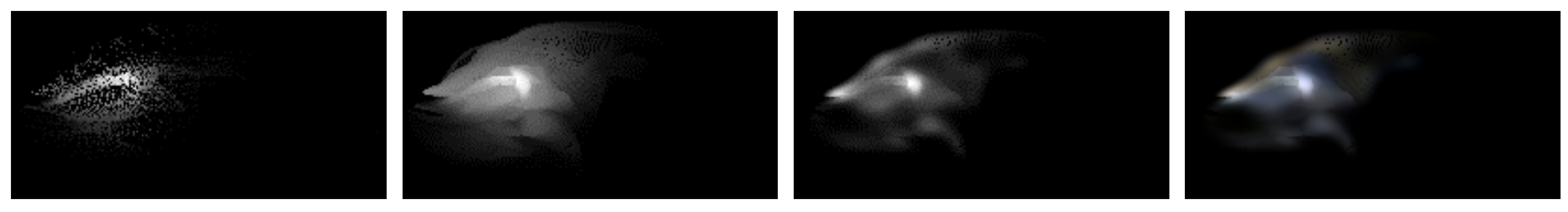}}\\
(a) & (b) & (c) & (d)\\
\end{tabular}
   \caption{Intermediate results of illumination estimation. (a) Traced environment map by forward warping; (b) Traced environment map by inverse warping;
(c) Map after deconvolution; (d) Final environment map after illumination color rescaling. }
\label{fig:project}
\end{figure}

As rigorously derived in \cite{ramamoorthi2001signal}, reflection can be expressed as the environment map convolved with the surface BRDF (bidirectional reflectance distribution function), e.g., the model in Equation~\ref{equation:Phong}. Therefore, if we trace the highlight component of a face back toward the scene, we obtain a convolved version of the environment map, where the convolution kernel is determined by the specular reflectance lobe. With surface normals computed using a single-image face reconstruction algorithm \cite{yang2011expression}, our method performs this tracing to recover an initial environment map, such as that exhibited in Figure~\ref{fig:project} (a).
%In this stage, we only use the grayscale intensity of the highlight channel, since its color is often different from the light source due to pixel intensity saturation in the input image. We will estimate the illumination color in Section~\ref{sec:albedoshading}.

Due to limited image resolution, the surface normals on a face are sparsely sampled, and an environment map obtained by directly tracing the highlight component would be sparse as well, as shown in Figure~\ref{fig:project} (a). To avoid this problem, we employ inverse image warping where for each pixel $p$ in the environment map, trace back to the face to get its corresponding normal $N_p$ and use the available face normals nearest to $N_p$ to interpolate a highlight value of $N_p$. In this way, we avoid the holes and overlaps caused by directly tracing (i.e., forward warping) highlights to the environment map. The result of this inverse warping is illustrated in Figure~\ref{fig:project} (b).

\subsection{Deconvolution by the specular lobe} \label{sec:deconv}
Next, we use the specular lobe to deconvolve the filtered environment map. This deconvolution is applied in the spherical domain, rather than in the spatial domain parameterized by latitude and longitude which would introduce geometric distortions.

Consider the deconvolution kernel $K_{x}$ centered at a point $\ve{x}=(\theta_x,\phi_y)$ on the environment map.
At a nearby point $\ve{y}=(\theta_y,\phi_y)$, the value of $K_x$ is
\begin{equation}
\begin{split}
K_x(\ve{y})=k_s^x (L_y \cdot L_x)^{\alpha_x}\label{equation:kernel}
\end{split}
\end{equation}
\noindent where $L_x$ and $L_y$ are 3D unit vectors that point from the sphere center toward $\ve{x}$ and $\ve{y}$, respectively.
The terms $\alpha_x$ and $k_s^x$ denote the surface roughness and specular albedo at $\ve{x}$.

%Basically, for pixels corresponding to normal values the same region, the deconvolution kernels are same. 

To determine $\alpha_x$ and $k_s^x$ for each pixel in the environment map, we use statistics from the MERL/ETH Skin Reflectance Database~\cite{weyrich2006analysis}. In these statistics, faces are categorized by skin type, and every face is divided into ten regions, each with its own mean specular albedo and roughness because of differences in skin properties, e.g., the forehead and nose being relatively more oily. 
Using the mean albedo and roughness value of each face region for the face's skin type\footnote{Skin type is determined by the closest mean albedo to the mean value of the face's albedo layer. Extraction of the face's albedo layer is described in Sec.~\ref{sec:albedoshading}.}, our method performs deconvolution by the Richardson-Lucy algorithm~\cite{lucy1974iterative,richardson1972bayesian}.
Figure~\ref{fig:project} (c) shows an environment map after deconvolution.

%-------------------------------------------------------------------------
\subsection{Rescaling illumination color}
\label{sec:albedoshading}
\begin{figure}[t]
\centering
\begin{tabular}
{>{\centering\arraybackslash}m{0.22\linewidth}
>{\centering\arraybackslash}m{0.2\linewidth}
>{\centering\arraybackslash}m{0.2\linewidth}
>{\centering\arraybackslash}m{0.22\linewidth}}
\multicolumn{4}{c}{\includegraphics[width=0.85\linewidth]{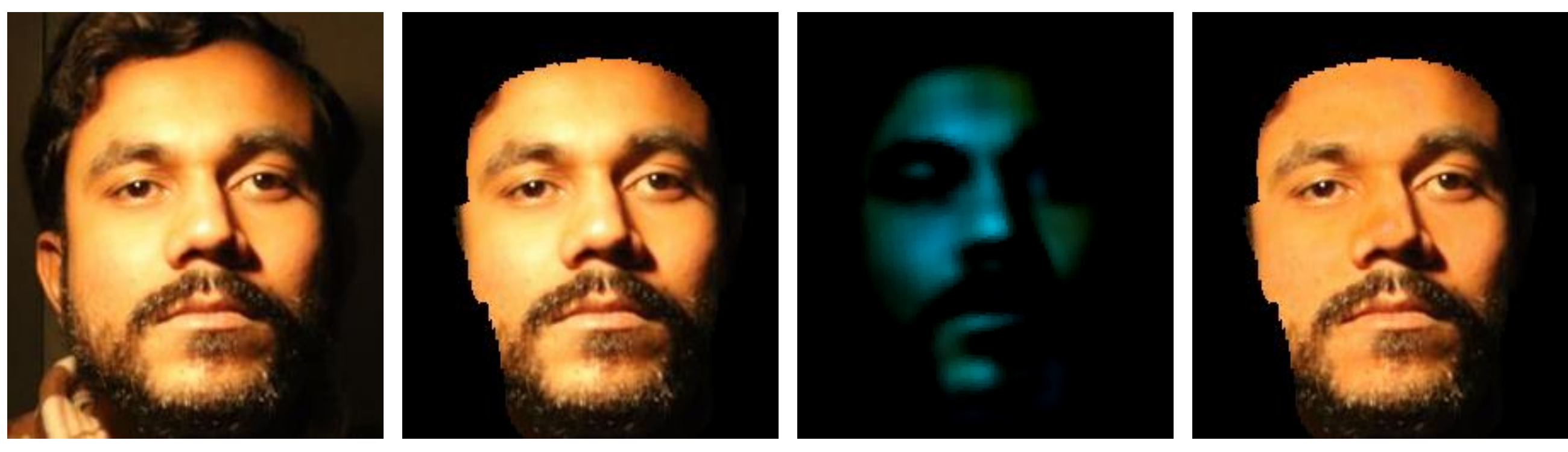}}\\
(a) & (b) & (c) & (d) \\
\end{tabular}
\caption{(a) Input photo; (b) Automatically cropped face region by landmarks \cite{zhu2012face} (network input); (c) predicted highlight layer (scaled by 2); (d) highlight removal result.}
\label{fig:blue}
\end{figure}

The brightness of highlight reflections often leads to saturated pixels, which have color values clipped at the maximum image intensity. As a result, the highlight intensity in these color channels may be underestimated.
This problem is illustrated in Figure \ref{fig:blue}, where the predicted highlight layer appears blue because the light energy in the red and green channels is not fully recorded in the input image.
%because the red and green channels are nearly saturated in the original input image, where the nonlinear camera response function compresses the highlight magnitude in these two channels. As a result, the separated highlight appears blue, though the lighting has strong red and green components. [Steve: our images are radiometrically calibrated]
To address this issue, we take advantage of diffuse shading, which is generally free of saturation and indicative of illumination color.

% Specifically, 
% \begin{equation}
% \begin{split}
% I=A\cdot S\label{equation:direct}
% \end{split}
% \end{equation}
% \noindent where an pixel-wise production is applied for the whole image. 
% By improving the reflectance model to consider diffuse and highlight layer separately, here we assume: 
% \begin{equation}
% \begin{split}
% I=A\cdot S_d+H\label{equation:direct}
% \end{split}
% \end{equation}
% \noindent where the image formation can be considered as the sum of highlight and pixel-wise production of albedo and diffuse shading. 
% \ping{why do you need this model $I = A\dot S + H$?? It seems you apply the intrinsic image decomposition on the diffuse channel, which is free of highlight already.}{\color{renjiao}{Yes our input is free of highlight already, I put this equation to show that original assumption is actually incorrect because specular components does not belong to the shading part. So I=A*s+H is a more precise model. If we combine this two CNN as showed in Figure. 6. We can separate 3 components correctly}}

%The diffuse shading is obtained through intrinsic image decomposition, which separates diffuse reflection into an albedo and a diffuse shading layer. 
Diffuse reflection (i.e., the diffuse layer) is the product of albedo and diffuse shading, and the diffuse shading can be extracted from the diffuse layer through intrinsic image decomposition.
To accomplish this decomposition, we finetune the intrinsic image network from~\cite{narihira2015direct} using synthetic face images to improve the network's effectiveness on faces. Specifically, 10,000 face images were synthesized from 50 face shapes randomly generated using the Basel Face Model \cite{paysan20093d}, three different skin tones, diffuse reflectance, and environment maps randomly selected from 100 indoor and 100 outdoor real HDR environment maps. 
%The ground truth diffuse shading layer $S_d$ is computed by $I_d=A\cdot S_d$, where $I_d$ are renderings of these diffuse faces, $A$ is the albedo layer (texture maps of the 150 faces). 
Adding this Albedo-Shading Net to our system as shown in Figure~\ref{fig:network2} (b) yields a highlight layer, albedo layer, and diffuse shading layer from an input face.

With the diffuse shading layer, we recolor the highlight layer $H$ extracted via Highlight-Net by rescaling its channels. When the blue channel is not saturated, its value is correct and the other channels are rescaled relative to it as 
\begin{equation}
\left[ H'(r), \; H'(g), \; H'(b) \right] = \left[ H(b)*c_d(r)/c_d(b), \; H(b)*c_d(g)/c_d(b), \; H(b) \right]
%\begin{split}
%& H'(b) = H(b), \\ %\hspace*{0,2cm}
%& H'(r) = H(b)*c_d(r)/c_d(b), \\
%& H'(g) = H(b)*c_d(g)/c_d(b)
\label{equation:recolor}
%\end{split}
\end{equation}
where $c_d$ is the diffuse shading chromaticity.
Rescaling can similarly be solved from the red or green channels if they are unsaturated. If all channels are saturated, we use the blue channel as it is likely to be the least underestimated based on common colors of illumination and skin.
After recoloring the highlight layer, we compute its corresponding environment map following the procedure in Sections~\ref{sec:init}-\ref{sec:deconv} to produce the final result, such as shown in Figure~\ref{fig:project} (d). 

%-------------------------------------------------------------------------

\subsection{Triangulating lights from multiple faces}
\label{sec:tri}
In a scene where the light sources are nearby, the incoming light distribution can vary significantly at different locations. 
%An environment map captures the illumination condition of distant lighting, e.g. in an outdoor scene. However, indoor scenes often involves nearby lights, which requires spatially variant environment maps\cite{gardner2017learning}\renjiao{In this paper they mentioned that indoor lighting cannot be considered as distant directional lighting. }\ping{cite a graphics paper talking about this kind of nearby indoor lighting effects}.  
%When there are multiple faces in an image, we can estimate an envrionment map centered at each face. 
An advantage of our non-parametric illumination model is that when there are multiple faces in an image, we can recover this spatially variant illumination by inferring the environment map at each face and using them to triangulate the 3D light source positions. 

As a simple scheme to demonstrate this idea, we first use a generic 3D face model (e.g., the Basel Face Model \cite{paysan20093d}) to solve for the 3D positions of each face in the camera's coordinate system, by matching 3D landmarks on the face model to 2D landmarks in the image using the method of \cite{zhu2012face}. Highlight-Net is then utilized to acquire the environment map at each of the faces. In the environment maps, strong light sources are detected as local maxima found through non-maximum suppression. To build correspondences among the lights detected from different faces, we first match them according to their colors. When there are multiple lights of the same color, their correspondence is determined by triangulating different combinations between two faces, with verification using a third face. In this way, the 3D light source positions can be recovered.
\section{Experiments}\label{sec:results}

\subsection{Training data}\label{sec:data}
For the pretraining of Highlight-Net, we use the Basel Face Model~\cite{paysan20093d} to randomly generate 50 3D faces.
For each face shape, we adjust the texture map to simulate three different skin tones. 
These 150 faces are then rendered under 200 different HDR environment maps, including 100 from indoor scenes and 100 from outdoor scenes. 
The diffuse and specular components are rendered separately, where a spatially uniform specular albedo is randomly generated between $[0,1]$. Some examples of these renderings are provided in the supplemental document. 
For training, we preprocessed each rendering by subtracting the mean image value and then normalizing to the range [0,1].

%\begin{figure}
%\centering
%\includegraphics[width=1 \linewidth]{Figures/hdrmaps.PNG}\
%   \caption{Real HDR environment maps used in rendering synthetic faces. Both indoor and outdoor environment maps are used for a variety of illuminations. }
%\label{fig:hdrmaps}
%\end{figure}

%\begin{figure}
%\centering
%\includegraphics[width=1 \linewidth]{Figures/trainingdata1-lowres.PNG}\
%\caption{Examples of rendered synthetic faces. From top to bottom, these rows show the diffuse, specular (scaled by 3 for visualization), and the combined layers. }
%\label{fig:syntheticdata}
%\end{figure}

In finetuning Highlight-Net, the image set for each celebrity undergoes a series of commonly-used preprocessing steps so that the faces are aligned, frontal, radiometrically calibrated, and under a consistent illumination color. For face frontalization, we apply the method in \cite{hassner2015effective}. We then identify facial landmarks \cite{zhu2012face} to crop and align these frontal faces. The cropped images are radiometrically calibrated by the method in \cite{li2017radiometric}, and their color histograms are matched by the built-in histogram transfer function in MATLAB~\cite{MATLAB} to reduce illumination color differences. We note that in each celebrity's set, images were manually removed if the face exhibits a strong expression or multiple lighting colors, since these cases often lead to inaccurate spatial alignment or poor illumination color matching. Some examples of these preprocessed images are presented in the supplementary material.
%Figure~\ref{fig:realface} shows some sample images of two celebrities. 
%, since they are collected from web captured by different and unknown cameras. At last, a color histogram transfer by built-in function in MATLAB\cite{MATLAB} is applied to align the color histogram among images for each celebrity, so after this, we can assume the illumination color in images are same. 

%\begin{figure}
%\centering
%\includegraphics[width=1 \linewidth]{Figures/trainingdata-lowres.png}\
%   \caption{Some selected preprocessed celebrity photos. }
%\label{fig:realface}
%\end{figure}

% under uniform color illumination, each pair of data (a pair contains an original image, and a diffuse image) are captured under a random lighting setting

\subsection{Evaluation of highlight removal}
\begin{figure*}[t]
\centering
\begin{tabular}
{>{\centering\arraybackslash}m{0.1\linewidth}
>{\centering\arraybackslash}m{0.1\linewidth}
>{\centering\arraybackslash}m{0.1\linewidth}
>{\centering\arraybackslash}m{0.1\linewidth}
>{\centering\arraybackslash}m{0.1\linewidth}
>{\centering\arraybackslash}m{0.1\linewidth}
>{\centering\arraybackslash}m{0.1\linewidth}
>{\centering\arraybackslash}m{0.1\linewidth}
>{\centering\arraybackslash}m{0.1\linewidth}}
\multicolumn{9}{c}{\includegraphics[width=0.95\linewidth]{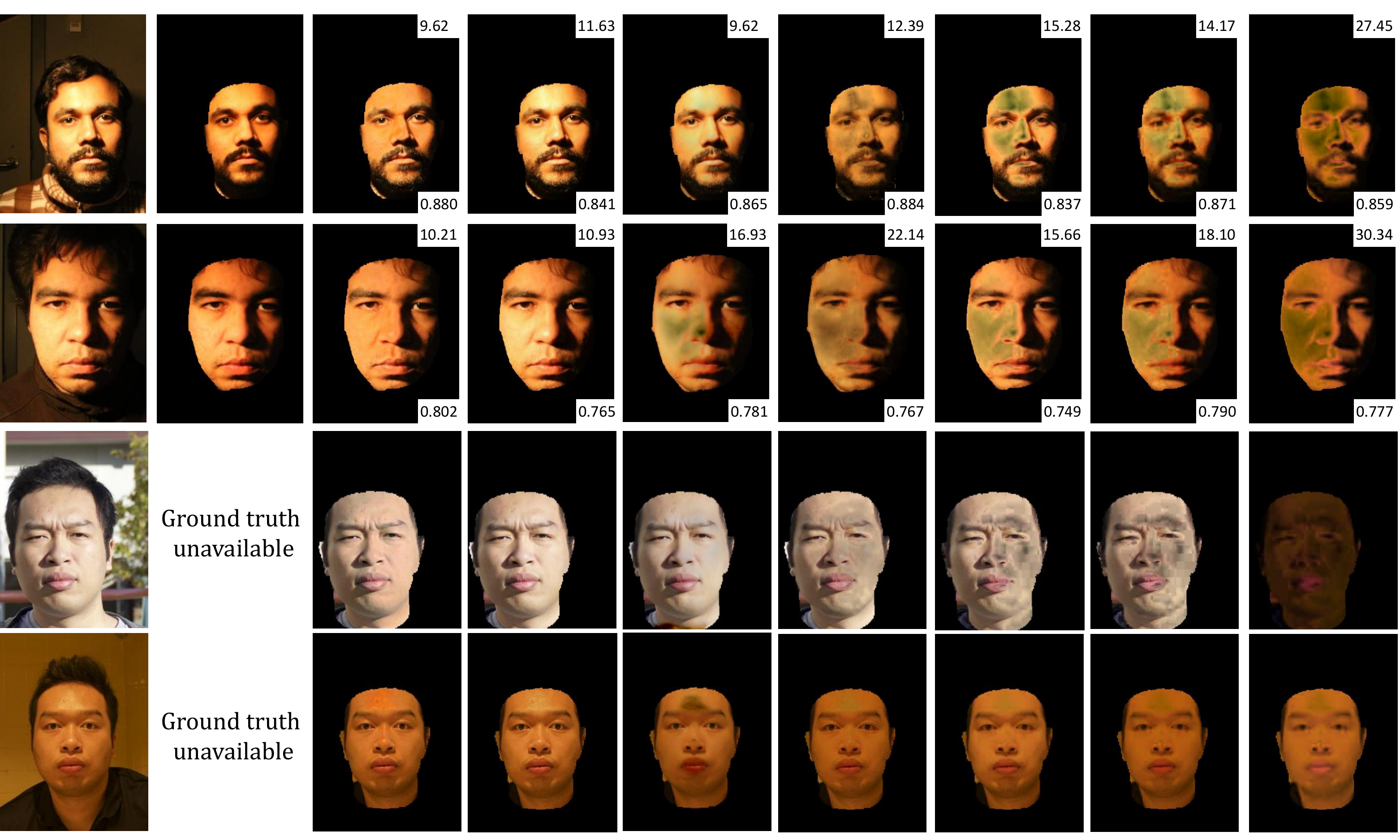}}\\
(a) & (b) & (c) & (d) & (e) & (f) & (g) & (h)&(i)\\
\end{tabular}
\caption{Highlight removal comparisons on laboratory images with ground truth and on natural images. Face regions are cropped out automatically by landmark detection \cite{zhu2012face}. (a) Input photo. (b) Ground truth captured by cross-polarization for lab data. (c-h) Highlight removal results by (c) our finetuned Highlight-Net, (d) Highlight-Net without finetuning, (e) \cite{shi2017learning}, (f) \cite{li2017specular}, (g) \cite{shen2013real}, (h) \cite{yang2010real}, and (i) \cite{tan2004separating}. For the lab images, RMSE values are given at the top-right, and SSIM \cite{wang2004image} (larger is better) at the bottom-right. }
\label{fig:highlightcompare}
\end{figure*}
To examine highlight extraction performance, we compare our highlight removal results to those of several previous techniques \cite{tan2004separating,shi2017learning,yang2010real,shen2013real,li2017specular} in Figure~\ref{fig:highlightcompare}. 
The first two rows show results on faces with known ground truth captured by cross-polarization under an indoor directional light. In order to show fair comparisons for both absolute intensity errors and structural similarities, we use both RMSE and SSIM\cite{wang2004image} as error/similarity metrics.
The last two rows are qualitative comparisons on natural outdoor and indoor illuminations, where ground truth is unavailable due to the difficulty of cross-polarization in general settings.
In all of these examples, our method outperforms the previous techniques, which generally have difficulty in dealing with the saturated pixels that commonly appear in highlight regions.
%While the method in \cite{li2017specular} is able to handle partial saturation (i.e., saturation of one color channel), it cannot deal with saturation of multiple channels, which frequently occurs in face images. 
We note that since most previous techniques are based on color analysis and the dichromatic reflection model \cite{shafer1985color}, they cannot process grayscale images, unlike our CNN-based method. For results on grayscale images and additional color images, please refer to the supplement.
The figure also illustrates the importance of training on real image data. Comparing our finetuning-based method in (c) to our method without finetuning in (d) and a CNN-based method trained on synthetic data \cite{shi2017learning} in (e) shows that training only on synthetic data is insufficient, and that our unsupervised approach for finetuning on real images substantially elevates the quality of highlight separation.
%The figure also compares the results of our method with and without finetuning, in (c) and (d). It is shown that training only on synthetic data is insufficient, and that our unsupervised approach for finetuning on real images substantially elevates the quality of highlight separation. \renjiao{Another CNN-based method\cite{shi2017learning} generates similar results with our model without finetuning on the first and the third data, which still have a lot of highlights left, further proved the effeciency of finetuning on real data.  }

Quantitative comparisons over 100 synthetic faces and 30 real faces are presented in Table~\ref{table:removal}. Error histograms and image results are shown in the supplement.

\begin{table}[t]
\centering 
\begin{tabular}{c|c c c c c c c| c c c c c c} 
\hline 
&
\multicolumn{6}{c}{Synthetic data}&&\multicolumn{6}{c}{Real data}\\
\hline
& Ours &\cite{shi2017learning}&\cite{li2017specular}&\cite{shen2013real}&\cite{yang2010real}&\cite{tan2004separating}&&Ours&\cite{shi2017learning}&\cite{li2017specular}&\cite{shen2013real}&\cite{yang2010real}&\cite{tan2004separating}\\ [0.5ex] 
\hline 
Mean RMSE&\bf{3.37}&4.15&5.35&6.75&8.08&28.00&&\bf{7.61}&8.93&10.34&10.51&11.74&19.60\\ 
Median RMSE&\bf{3.41}&3.54&4.68&6.41&7.82&29.50&&\bf{6.75}&8.71&10.54&9.76&11.53&22.96\\ 
Mean SSIM&\bf{0.94}&\bf{0.94}&0.92&0.91&0.91&0.87&&0.89&0.89&\bf{0.90}&0.86&0.88&0.88 \\ 
Median SSIM&\bf{0.95}&0.94&0.92&0.91&0.91&0.87&&0.90&0.90&\bf{0.91}&0.88&0.90&0.89\\  
\hline
\end{tabular}
\caption{Quantitative highlight removal evaluation. } 
\label{table:removal} 
\end{table}

%\subsection{Evaluation of illumination estimation}
%\begin{figure*}
%\centering
%\begin{tabular}
%{>{\centering\arraybackslash}m{0.11\linewidth}
%>{\centering\arraybackslash}m{0.12\linewidth}
%>{\centering\arraybackslash}m{0.13\linewidth}
%>{\centering\arraybackslash}m{0.14\linewidth}
%>{\centering\arraybackslash}m{0.14\linewidth}
%>{\centering\arraybackslash}m{0.14\linewidth}
%>{\centering\arraybackslash}m{0.14\linewidth}}
%\multicolumn{7}{c}{\includegraphics[width=0.97\linewidth]{Figures/figures-envmaps-nov14-lowres.png}}\\
% (a) & (b) & (c) & (d) & (e) & (f) & (g)\\
%\end{tabular}
%   \caption{Comparison of environment maps estimated by different methods. (a) Input face photo. (b) Ground truth captured using a mirrored sphere. (c-g) Environment maps from (c) Our method, (d) \cite{hold2016deep}, (e) \cite{gardner2017learning}, (f) \cite{lombardi2016reflectance}, and (g) \cite{knorr2014real} (spherical harmonics representation). \renjiao{to-do: move to the supplementary material, re-expose the environment maps and add more synthetic data with more ambient lighting.}}
%\label{fig:envmaps}
%\end{figure*}

\begin{table}[t]
\centering 
\begin{tabular}{c|c c c c c c c| c c c c c c} 
\hline
&
\multicolumn{6}{c}{Diffuse Bunny}&&\multicolumn{5}{c}{Glossy Bunny}\\
\hline 
Relighting RMSE & Ours & \cite{hold2016deep} & \cite{gardner2017learning} & \cite{lombardi2016reflectance} & \multicolumn{2}{l}{\cite{knorr2014real}}&&Ours & \cite{hold2016deep} & \cite{gardner2017learning} & \cite{lombardi2016reflectance} & \cite{knorr2014real}\\ [0.5ex] 
\hline 
Mean (outdoor) & \bf{10.78} & 18.13 & $\backslash$ & 21.20&  \multicolumn{2}{l}{17.77}&&\bf{11.02}  & 18.28 & $\backslash$ &21.63 &  18.28 \\ 
Median (outdoor)& \bf{9.38} & 17.03 & $\backslash$ & 19.95&  \multicolumn{2}{l}{15.91}&& \bf{9.74} & 17.67 & $\backslash$ &20.49& 16.30  \\ 
Mean (indoor) & \bf{13.18} &  $\backslash$ &29.25&25.40&  \multicolumn{2}{l}{20.52}&&\bf{13.69}  &  $\backslash$& 29.71 & 25.92  &21.01  \\ 
Median (indoor)& \bf{11.68} & $\backslash$ &25.99& 25.38&  \multicolumn{2}{l}{19.22}&& \bf{11.98} & $\backslash$ & 26.53 & 25.91& 19.75  \\  
\hline
\end{tabular}
\caption{Illumination estimation on synthetic data.} 
\label{table:synthetic} 
\end{table}
%\vspace*{-1.5cm}

%\begin{table}
%\centering 
%\begin{tabular}{c|c c c c c c c| c c c c c c} 
%\hline 
%Normalized RMSE & Ours & \cite{hold2016deep} & \cite{gardner2017learning} & \cite{lombardi2016reflectance} & \cite{knorr2014real}&\\ [0.5ex] 
%\hline 
%Mean &\bf{0.058}&0.126&0.086&0.092&0.282\\ 
%Median &\bf{0.042}&0.124&\bf{0.042}&0.082&0.275\\ 
%\hline 
%\end{tabular}
%\caption{Illumination estimation on real data.} 
%\label{table:real} 
%\end{table}

\begin{figure*}[t]
\centering
\begin{tabular}
{>{\centering\arraybackslash}m{0.2\linewidth}
>{\centering\arraybackslash}m{0.17\linewidth}
>{\centering\arraybackslash}m{0.18\linewidth}
>{\centering\arraybackslash}m{0.17\linewidth}
>{\centering\arraybackslash}m{0.17\linewidth}}
\multicolumn{5}{c}{\includegraphics[width=0.9\linewidth]{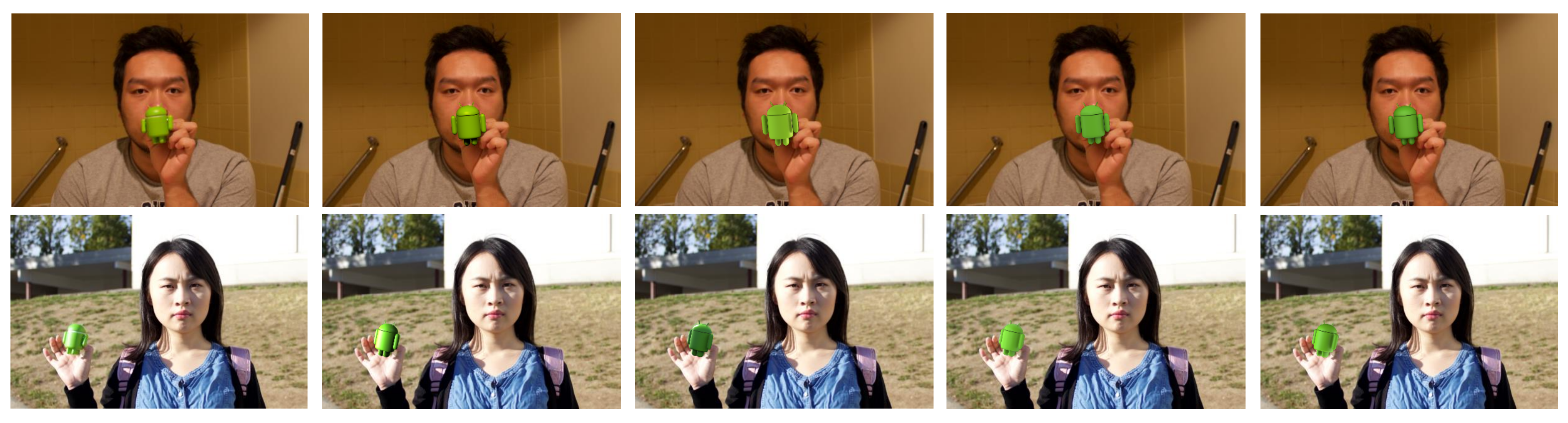}}\\
(a) & (b) & (c) & (d) & (e)\\
\end{tabular}
   \caption{Virtual object insertion results for indoor (first row) and outdoor (second row) scenes. (a) Photos with real object. Object insertion by (b) our method, (c) \cite{gardner2017learning} for the first row and \cite{hold2016deep} for the second row, (d) \cite{lombardi2016reflectance}, (e) \cite{knorr2014real}. More results in the supplement. }
\label{fig:outdoor}
\end{figure*}
\begin{figure}
\centering
\includegraphics[width=0.88\linewidth]{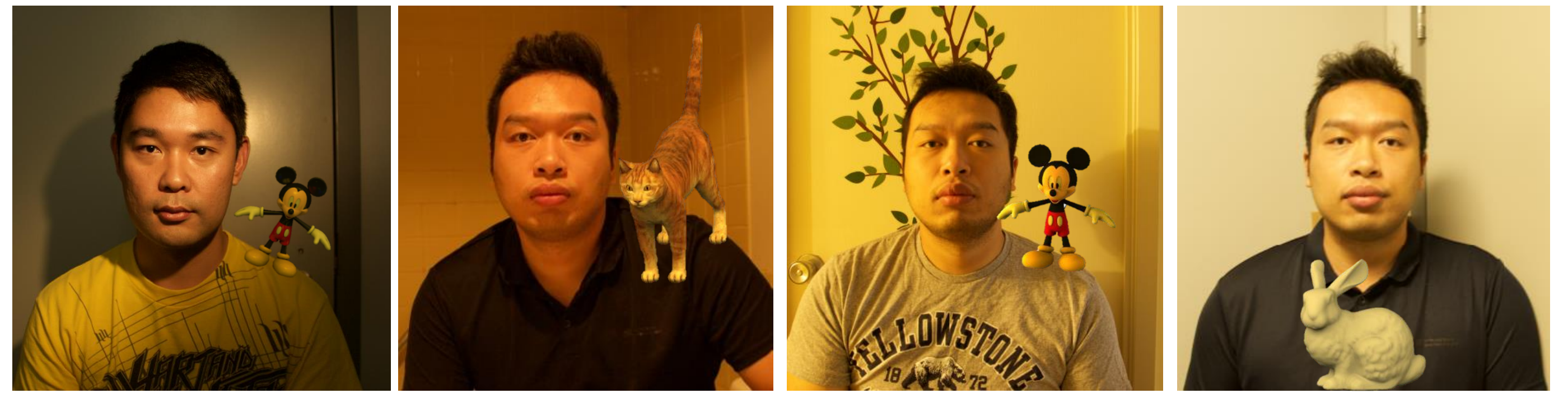}\
   \caption{Object insertion results by our method. }
\label{fig:insertion}
\end{figure}

\subsection{Evaluation of illumination estimation}
Following \cite{hold2016deep}, we evaluate illumination estimation by examining the relighting errors of a Stanford bunny under predicted environment maps and the ground truth. The lighting estimation is performed on synthetic faces rendered into captured outdoor and indoor scenes and their recorded HDR environment maps. Results are computed for both a diffuse and a glossy Stanford bunny (see the supplement for rendering parameters, visualization of rendered bunnies, and estimated environment maps). The comparison methods include the following: our implementation of \cite{knorr2014real} which uses a face to recover spherical harmonics (SH) lighting up to second order under the assumption that the face is diffuse; downloaded code for \cite{lombardi2016reflectance} which estimates illumination and reflectance given known surface normals that we estimate using \cite{yang2011expression}; online demo code for \cite{hold2016deep} which is designed for outdoor images; and author-provided results for \cite{gardner2017learning} which is intended for indoor images.

The relighting errors are presented in Table \ref{table:synthetic}. Except for \cite{hold2016deep} and \cite{gardner2017learning}, the errors were computed for 500 environment maps estimated from five synthetic faces under 100 real HDR environment maps (50 indoor and 50 outdoor). Since \cite{hold2016deep} and \cite{gardner2017learning} are respectively for outdoor and indoor scenes and are not trained on faces, their results are each computed from LDR crops from the center of the 50 indoor/outdoor environment maps. 
We found \cite{hold2016deep} and \cite{gardner2017learning} to be
generally less precise in estimating light source directions, especially when light sources are out-of-view in the input crops, but they still provide reasonable approximations. 
For \cite{gardner2017learning}, the estimates of high frequency lighting become less precise when the indoor environment is more complicated.
The experiments indicate that \cite{lombardi2016reflectance} may be relatively sensitive to surface textures and imprecise geometry in comparison to our method, which is purposely designed to deal with faces. For the Spherical Harmonics representation \cite{knorr2014real}, estimates of a low-order SH model are seen to lack detail, and the estimated face albedo incorporates the illumination color, which leads to environment maps that are mostly white (see supplement for examples). Overall, the results indicate that our method provides the closest estimates to the ground truth. For a comparison of environment map estimation errors in real scenes, please refer to the supplement.

%\cite{hold2016deep} also estimate a close approximation, but they have some difficulties estimating the illumination color. For \cite{gardner2017learning}, the color of the ambient lighting are estimated well, but the high frequency part is less precise in color and directions when indoor environment become complicated, \cite{lombardi2016reflectance} and \cite{knorr2014real} estimate a low frequency lighting, which does not include a high frequency part.

%Images and their HDR environment maps are captured for 15 real scenes (7 indoor and 8 outdoor), with faces excluded and a larger field of view for \cite{hold2016deep} and \cite{gardner2017learning} as shown in the supplement. Errors with respect to the captured ground truths are presented in Table \ref{table:real} in terms of RMSE normalized by the difference of the maximum and minimum intensity of the estimated environment map, which is commonly used to facilitate comparison between data with different scales, like the intensity scaling factor of environment maps.

We additionally conducted comparisons on virtual object insertion using estimated illumination, as shown in Figure~\ref{fig:outdoor} and in the supplement. To aid in verification, we also show images that contain the actual physical object (an Android robot). 
In some cases such as the bottom of (c), lighting from the side is estimated as coming from farther behind, resulting in a shadowed appearance. Additional object insertion results are shown in Figure~\ref{fig:insertion}.

%as input of method in \cite{} to estimate the illumination. For indoor data, we also captured a background image corresponding to each data, as input of the method in \cite{gardner2017learning}. 

\subsection{Demonstration of light source triangulation}

\begin{figure}
\centering
\begin{tabular}
{>{\centering\arraybackslash}m{0.35\linewidth}
>{\centering\arraybackslash}m{0.11\linewidth}
>{\centering\arraybackslash}m{0.25\linewidth}
>{\centering\arraybackslash}m{0.25\linewidth}}
\multicolumn{4}{c}{\includegraphics[width=0.95\linewidth]{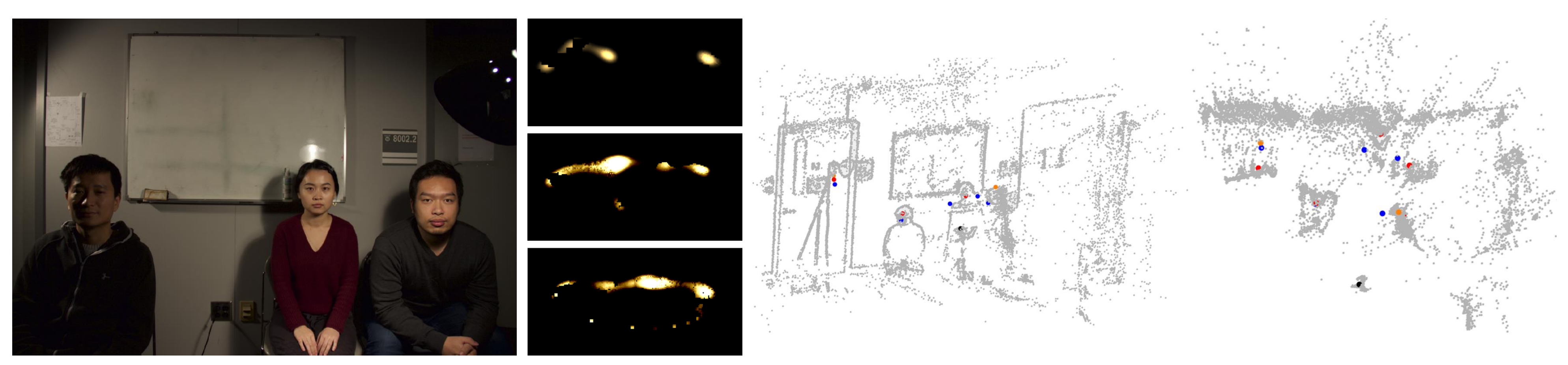}}\\
(a) & (b) & (c) & (d)\\
\end{tabular}
   \caption{(a) Input image with multiple faces; (b) their estimated environment maps (top to bottom are for faces from left to right); estimated 3D positions from (c) side view and (d) top view. Black dot: camera. Red dots: ground truth of faces and lights. Blue dots: estimated faces and lights. Orange dots: estimated lights using ground truth of face positions.}
\label{fig:3Dimage}
\end{figure}

Using the simple scheme described in Section~\ref{sec:tri}, we demonstrate the triangulation of two local light sources from an image with three faces, shown in Figure~\ref{fig:3Dimage} (a). The estimated environment maps from the three faces are shown in Figure~\ref{fig:3Dimage} (b). We triangulate the point lights from two of them, while using the third for validation.
In order to provide a quantitative evaluation, we use the DSO SLAM system \cite{engel2017direct} to reconstruct the scene, including the faces and light sources. We manually mark the reconstructed faces and light sources in the 3D point clouds as ground truth. 
As shown in Figure~\ref{fig:3Dimage} (c-d), the results of our method are close to this ground truth. The position errors are 0.19m, 0.44m and 0.29m for the faces from left to right, and 0.41m and 0.51m for the two lamps respectively. If the ground truth face positions are used, the position errors of the lamps are reduced to 0.20m and 0.49m, respectively. 

%\ping{why the reduction on the second error is so much smaller??}{\color{renjiao}{I think it is because our estimation of face positions of the left face is almost correct, but for the center and the right faces are not that correct. So while using the ground truth of face positions, the light source in front of the center and the right person are more accurate, but the light at the side are not improved a lot. I think it means the error of the light at the side is come from the environment maps, the error of the first light comes from the face positions. }}

%For a input image showed in \ref{fig:3Dimage}, there are more than one face in the scene, we can get location illumination of each face. Similar to many indoor scenes, due to relative positions of faces and light sources, the environment maps are strongly related with positions. 

\section{Conclusion}

%\begin{figure}
%\centering
%\includegraphics[width= 1\linewidth]{Figures/3dposition3.PNG}\
%   \caption{Estimated 3D positions. Left: side view. Right: top view. Black dot: camera. Red dots: ground truth of faces and lights. Blue dots: estimated faces and lights. Orange dots: estimated lights using ground truth of face positions.}
%\label{fig:3dposition}
%\end{figure}

We proposed a system for non-parametric illumination estimation based on an unsupervised finetuning approach for extracting highlight reflections from faces. In future work, we plan to examine more sophisticated schemes for recovering spatially variant illumination from the environment maps of multiple faces in an image. 
%Additionally, we will investigate methods for estimating the specular lobe parameters of a face directly from the image, instead of relying on reflectance statistics. 
Using faces as lighting probes provides us with a better understanding of the surrounding environment not viewed by the camera, which can benefit a variety of vision applications.

%{\color{renjiao}{In this paper, we proposed a unsupervised scheme to train a CNN to extract highlight layer from face images, the extracted highlight layer are then used to estimate a non-parametric illumination model. The highlight removal results outperform previous methods, and it can be used on grayscale images which previous methods cannot. Furthermore, illumination estimation can recover high frequency illumination, and spatial invariant illumination color. Using faces as lighting probes, with helps of prior knowledge on geometry, showed a high accuracy on recovering illumination, which can benefit many application for Augmented Reality. }}
\subsubsection{Acknowledgments. } We want to thank our colleagues from Gruvi Lab for capturing the face photos. This work is supported by Canada NSERC Discovery Grant 611664. Renjiao Yi is supported by scholarship from China Scholarship Council.

\bibliographystyle{splncs04}
\bibliography{egbib}
\clearpage
\appendix
\chapter*{Supplementary Material: \\Faces as Lighting Probes via Unsupervised Deep Highlight Extraction}
%\renewcommand{\theequation}{A.\arabicc}
%\theequation{Highlight-Net structure}
%\renewcommand{\section}{\arabic {section}}
\section{Highlight-Net structure}

\begin{figure}
\centering
\includegraphics[width=0.7 \linewidth]{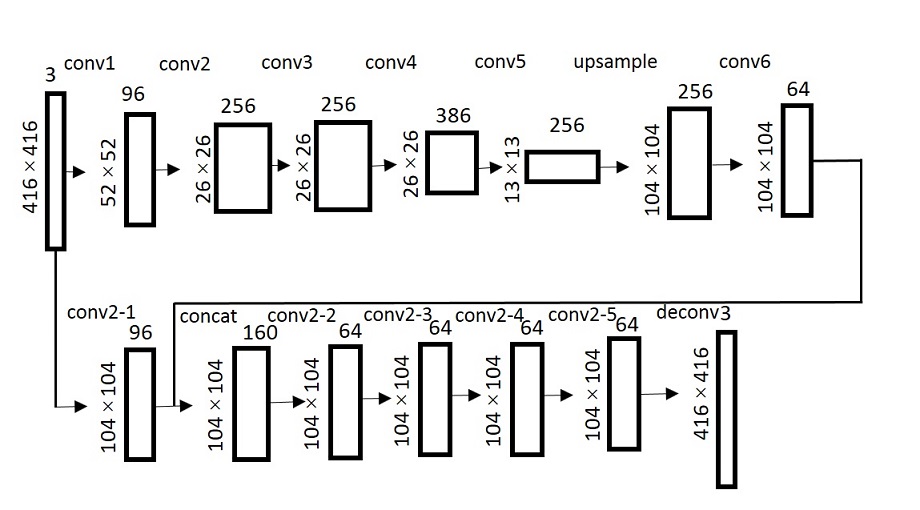}\
   \caption{Structure of Highlight-Net.}
\label{fig:structure}
\end{figure}

As mentioned in the paper, the structure of Highlight-Net is adopted from \cite{narihira2015direct}. The network structure is exhibited in Figure \ref{fig:structure}. 

\section{Training data for pretraining and finetuning}

\begin{figure}
\centering
\begin{tabular}
{>{\centering\arraybackslash}m{0.5\linewidth}
>{\centering\arraybackslash}m{0.5\linewidth}}
\multicolumn{2}{c}{\includegraphics[width=\linewidth]{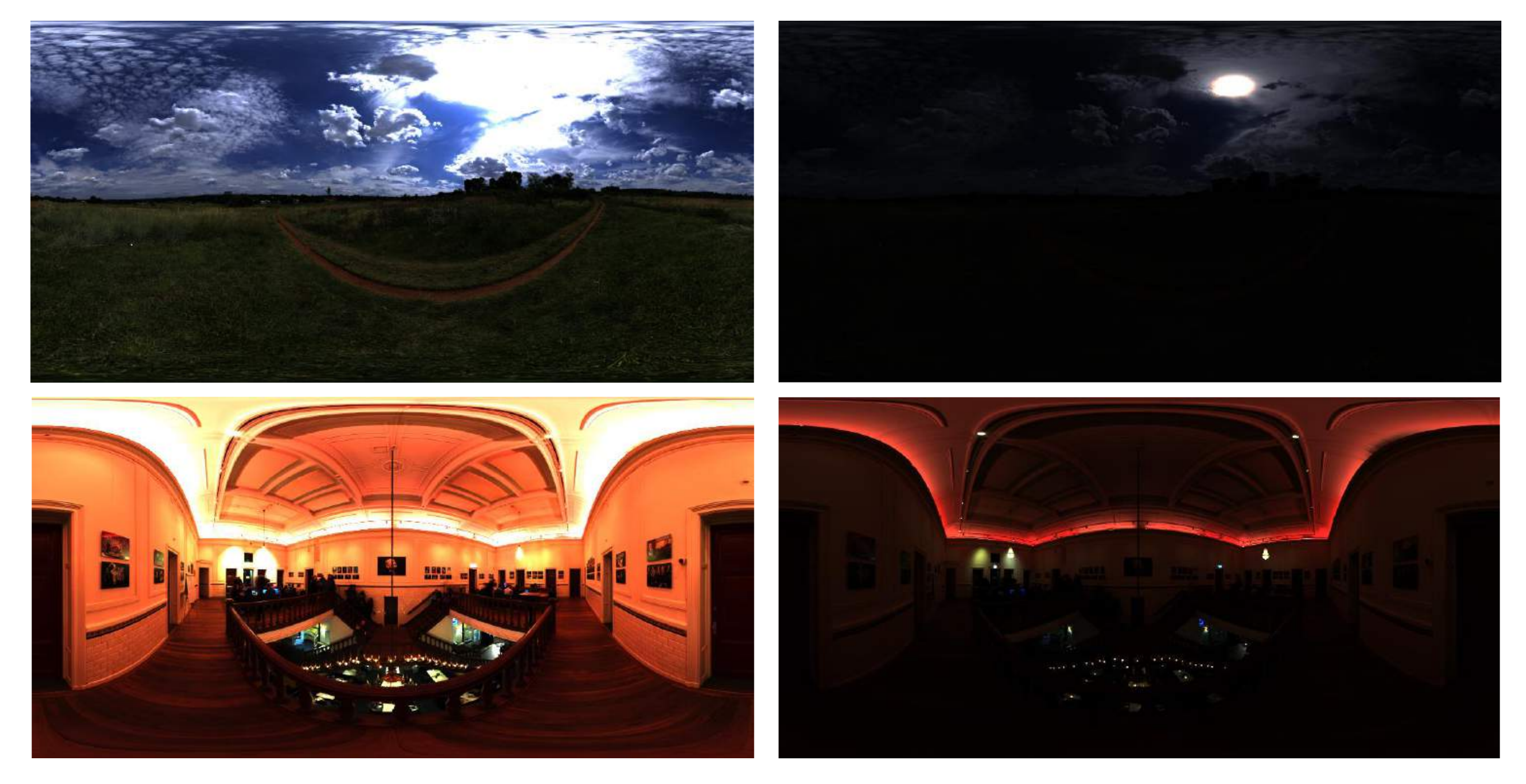}}\\
scaling factor = 1 & scaling factor = $10^{-1}$ \\
\end{tabular}
   \caption{Two example HDR environment maps for outdoor (top) and indoor (bottom) scenes. Shown at different scalng factors for better visualization of the high dynamic range.}
\label{fig:hdrmaps}
\end{figure}

\begin{figure}
\centering
\includegraphics[width=1 \linewidth]{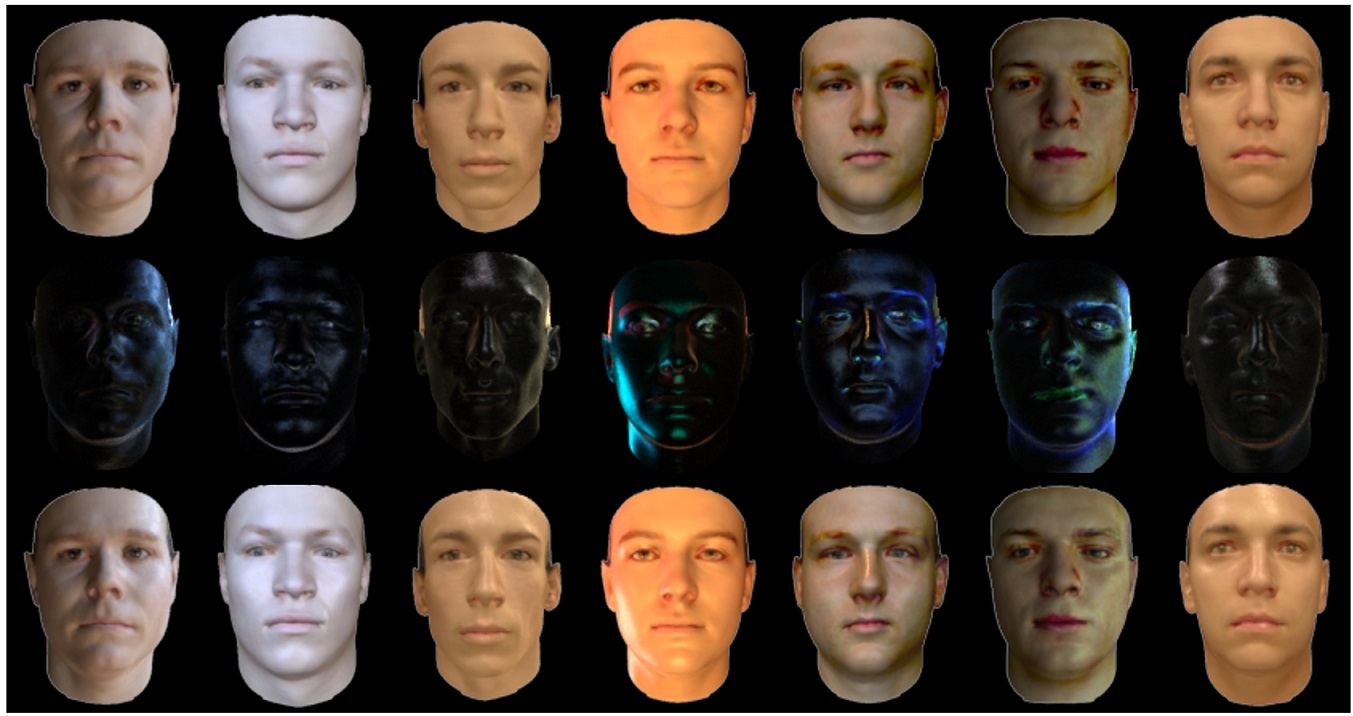}\
   \caption{Examples of rendered synthetic faces. The top row shows rendered diffuse components; the middle row displays rendered specular components; and the bottom row are composite renderings that combine the diffuse and specular layers.}
\label{fig:syntheticdata}
\end{figure}

As mentioned in the paper, for pretraining we rendered synthetic faces under real HDR environment maps, consisting of 100 indoor scenes and 100 outdoor scenes. Two examples of the environment maps are shown in Figure \ref{fig:hdrmaps}. Examples of rendered diffuse and specular layers, as well as the composite renderings, are displayed in Figure \ref{fig:syntheticdata}. 

\begin{figure}
\centering
\includegraphics[width=1 \linewidth]{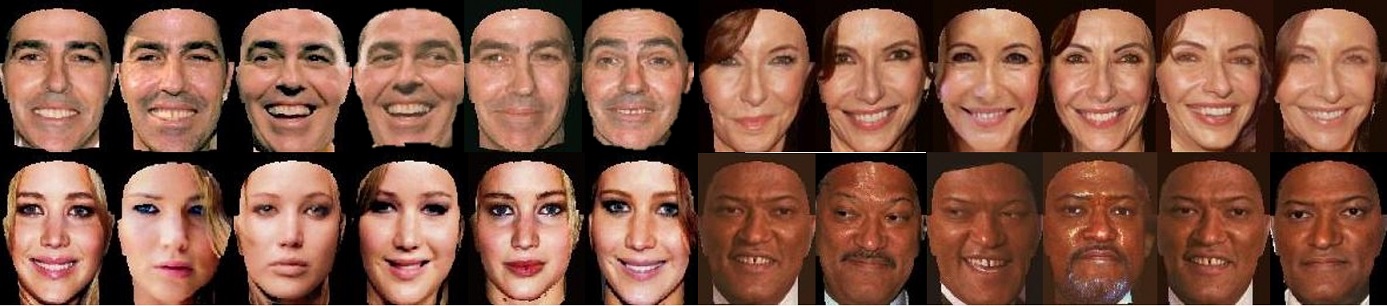}\
   \caption{Examples of preprocessed photos for four celebrities.}
\label{fig:realdata}
\end{figure}

In finetuning, as mentioned in Section 6.1 of the main text, images from the MS-celeb-1M dataset \cite{guo2016msceleb} are preprocessed by cropping and aligning based on landmarks detected by \cite{zhu2012face}, radiometric calibration by \cite{li2017radiometric}, and color histogram transfer to align illumination colors for each celebrity, performed by transferring the color histograms from one photo of each celebrity to the other photos of the same celebrity. Examples of preprocessed data are shown in Figure \ref{fig:realdata}.

\section{Additional results on highlight removal}

%Figure \ref{fig:highlightsaturated} exhibits highlight removal results of our method on images with strongly saturated regions, which are problematic for previous techniques. It can be seen that the results from Highlight-Net look natural and similar to the ground truth diffuse layer captured by cross-polarization. 

\begin{figure}
\centering
\includegraphics[width=1 \linewidth]{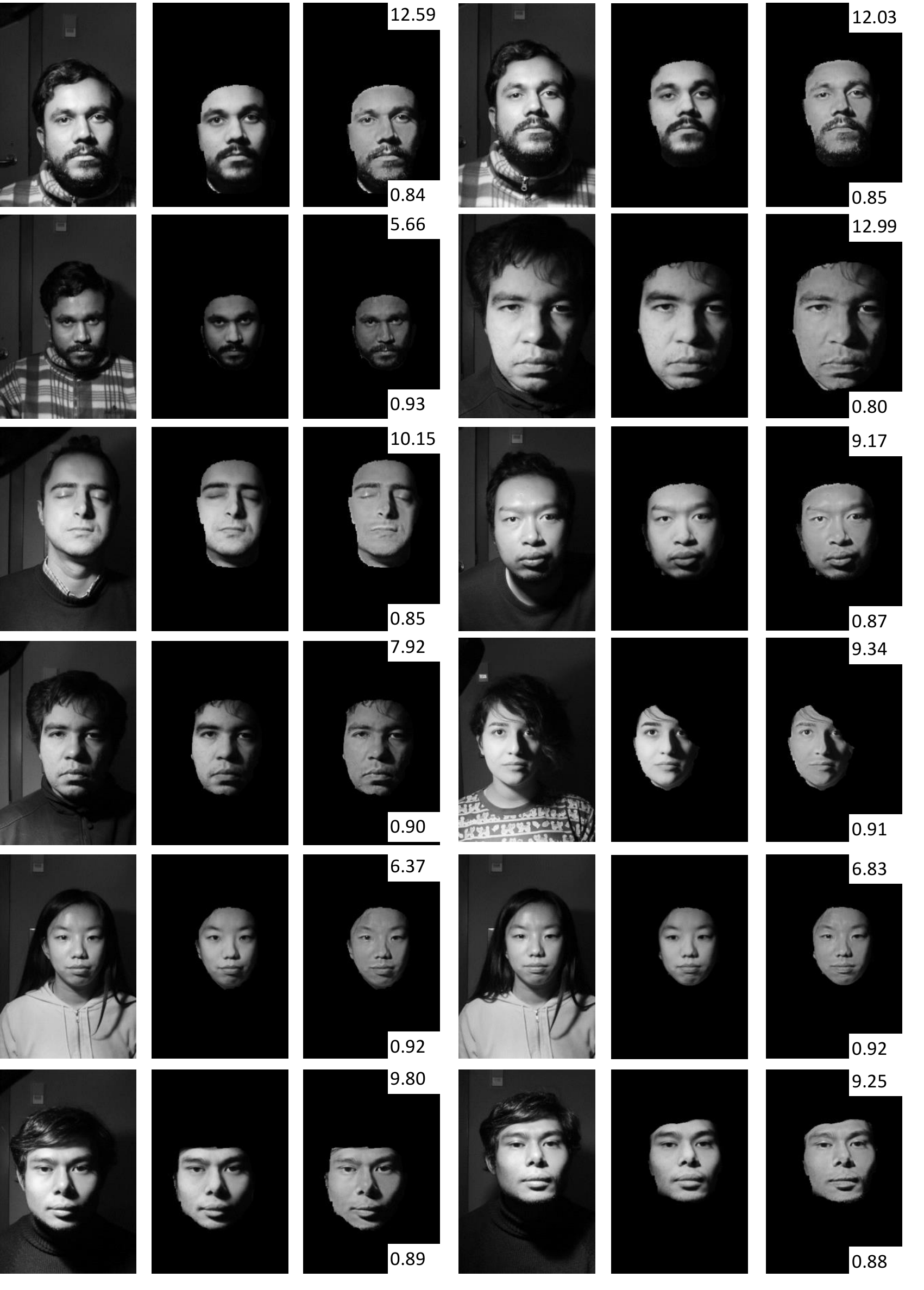}\
   \caption{Highlight removal in grayscale images by Highlight-Net. For each example, the input image, ground truth diffuse image, and our result are displayed from left to right. RMSE is given at the top-right of our results, and SSIM are shown at the bottom-right.  }
\label{fig:grayresult1}
\end{figure}

\begin{figure}
\centering
\includegraphics[width=1 \linewidth]{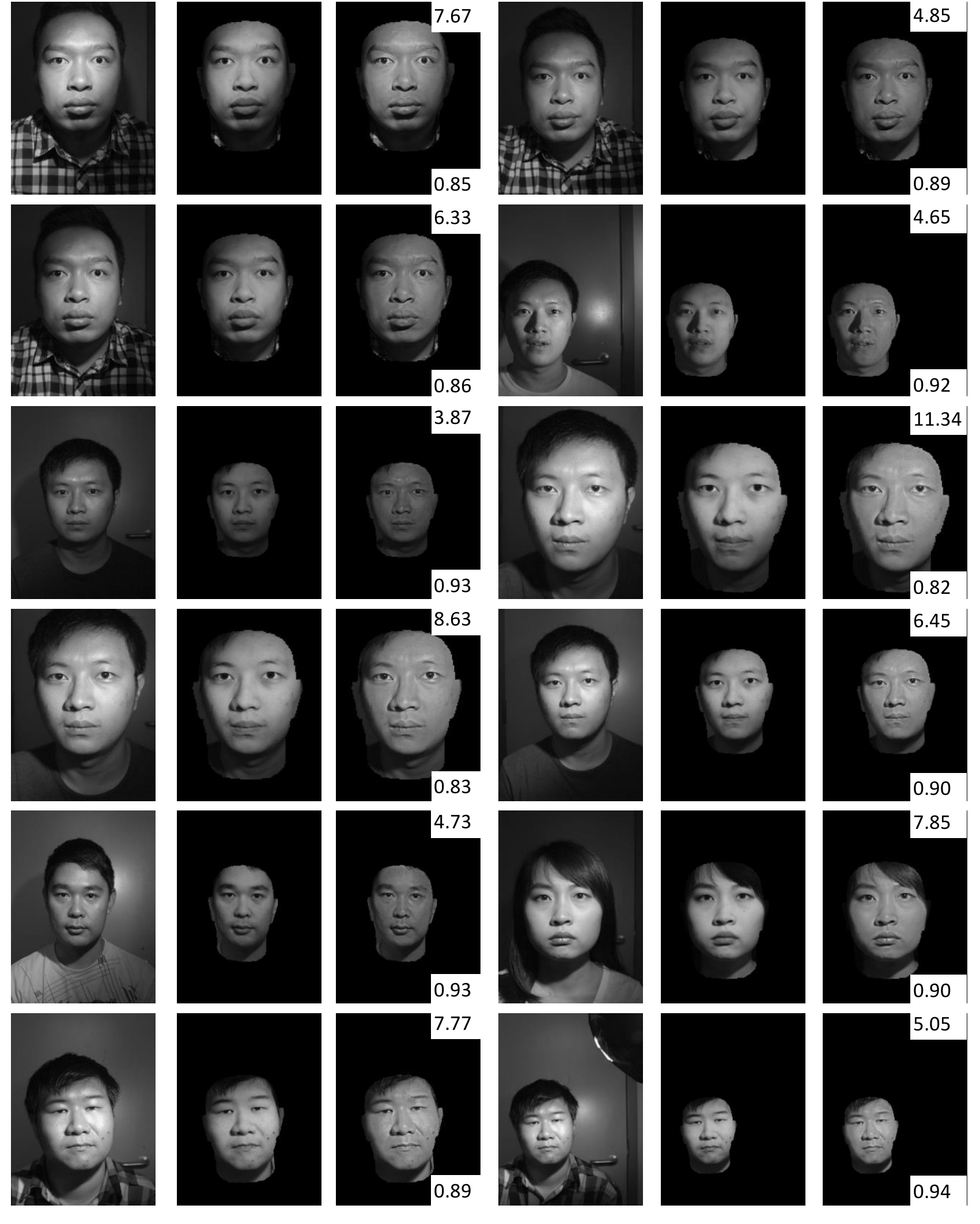}\
   \caption{Highlight removal in grayscale images by Highlight-Net. For each example, the input image, ground truth diffuse image, and our result are displayed from left to right. RMSE is given at the top-right of our results, and SSIM are shown at the bottom-right.  }
\label{fig:grayresult2}
\end{figure}

Highlight-Net can work for grayscale photos, unlike most previous methods which are based on color analysis. For a grayscale image, we input a color image whose three channels are equal to those of the grayscale photo. Then with the output, we average the values of the three channels to obtain the result. Although we do not train Highlight-Net on grayscale images, it nevertheless can produce reasonable results as shown in Figure~\ref{fig:grayresult1} and Figure~\ref{fig:grayresult2}, where RMSE and SSIM \cite{wang2004image} are marked in the figures for each example. RMSE represents absolute intensity errors, while SSIM measures structural similarity. Over all of the 30 real images that we captured together with cross-polarized ground truth, the mean SSIM and RMSE are 0.891 and 8.13, respectively. Like for the quantitative evaluation on color images, the RMSE and SSIM are computed on the highlight layer, because input images and ground truth matte images may already have a high structural similarity.
%\ping{Do we have the ground truth for those grayscale images? If so, we can further compute RMSE and SSIM.} 
Finetuning the net with grayscale training examples should lead to improvements in performance. 

\begin{figure}[t]
\centering
%\begin{tabular}
%{>{\centering\arraybackslash}m{0.3\linewidth}
%>{\centering\arraybackslash}m{0.23\linewidth}
%>{\centering\arraybackslash}m{0.23\linewidth}
%>{\centering\arraybackslash}m{0.24\linewidth}}
%\multicolumn{4}{c}{\includegraphics[width=0.95\linewidth]{Figures/highlighthist-new2.pdf}}\\
%(a)  & (b)  & (c)  & (d) \\
%\multicolumn{4}{c}{
\includegraphics[width=0.85\linewidth]{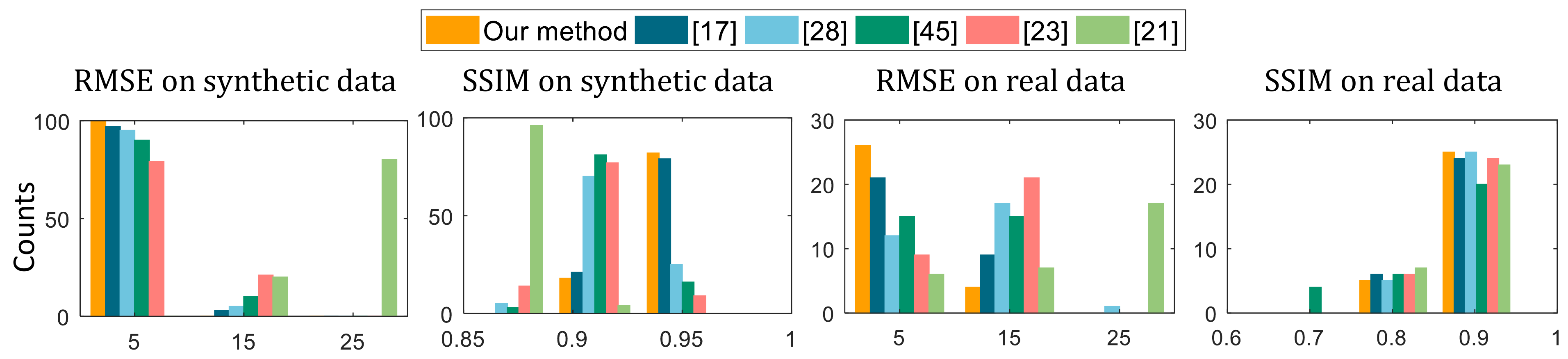}
%}\\
%(a)  & (b)  & (c)  & (d) \\
%\end{tabular}
   \caption{Quantitative comparisons on highlight removal for 100 synthetic faces and 30 real faces in terms of RMSE and SSIM histograms (larger SSIM is better). 
   %\renjiao{add \cite{li2017specular} in (a-b)} \ping{better to mark RMSE and SSIM into the chart directly (e.g. as the title of the chart)}
   }
\label{fig:highlighthis}
\end{figure}

\begin{figure}
\centering
\begin{tabular}
{>{\centering\arraybackslash}m{0.13\linewidth}
>{\centering\arraybackslash}m{0.11\linewidth}
>{\centering\arraybackslash}m{0.11\linewidth}
>{\centering\arraybackslash}m{0.1\linewidth}
>{\centering\arraybackslash}m{0.1\linewidth}
>{\centering\arraybackslash}m{0.1\linewidth}
>{\centering\arraybackslash}m{0.1\linewidth}
>{\centering\arraybackslash}m{0.1\linewidth}
>{\centering\arraybackslash}m{0.1\linewidth}}
\multicolumn{9}{c}{\includegraphics[width=0.99\linewidth]{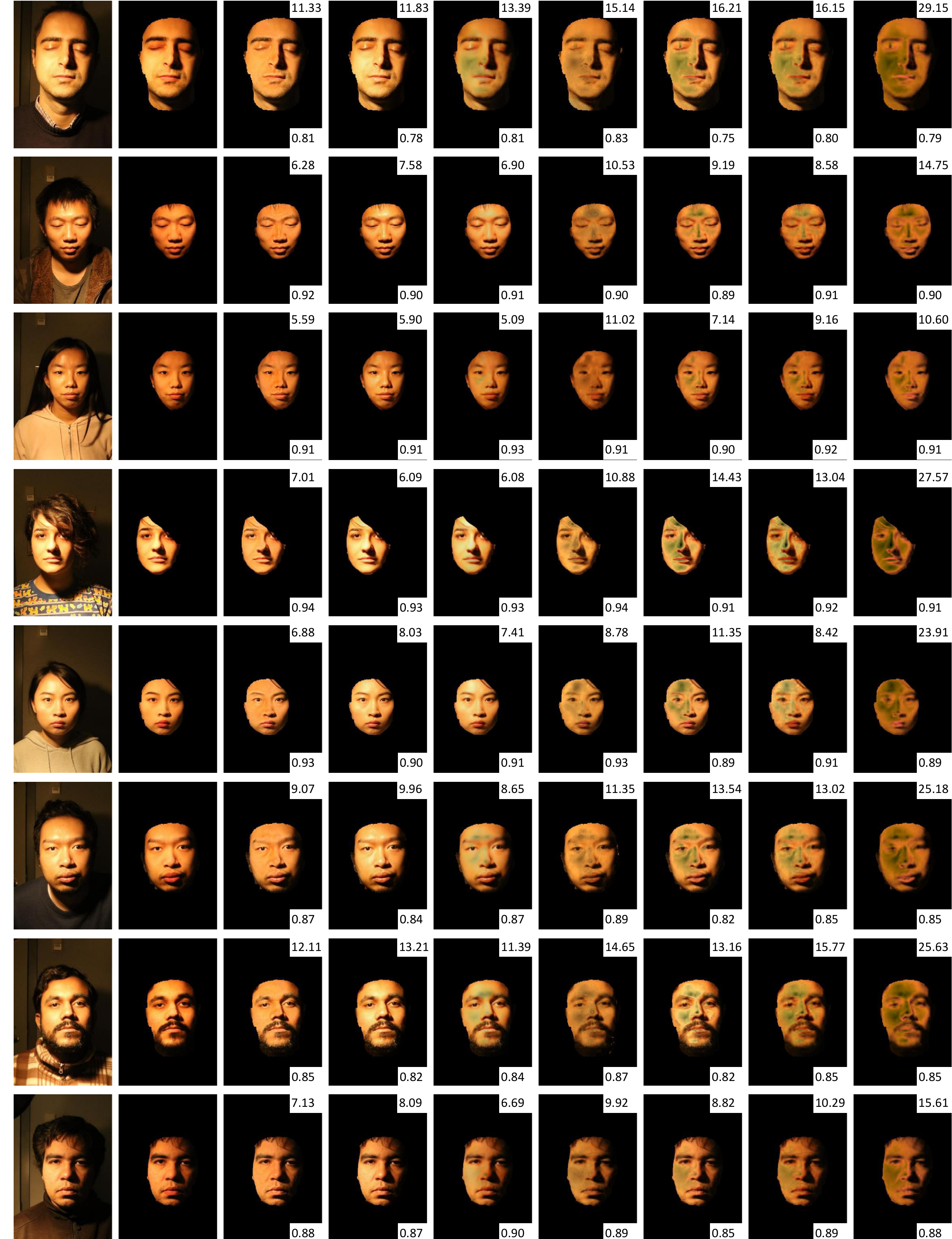}}\\
(a) & (b) & (c) & (d) & (e) & (f) & (g) & (h)&(i)\\
\end{tabular}
\caption{Additional highlight removal comparisons on laboratory images with ground truth. Face regions are cropped out automatically by landmark detection \cite{zhu2012face}. (a) Input photo. (b) Ground truth captured by cross-polarization for lab data. (c-h) Highlight removal results by (c) our finetuned Highlight-Net, (d) Highlight-Net without finetuning, (e) \cite{shi2017learning}, (f) \cite{li2017specular}, (g) \cite{shen2013real}, (h) \cite{yang2010real}, and (i) \cite{tan2004separating}. RMSE values are given at the top-right, and SSIM at the bottom-right. RMSE and SSIM are computed on highlight layers. }
\label{fig:highlightcompare1}
\end{figure}

\begin{figure}
\centering
\begin{tabular}
{>{\centering\arraybackslash}m{0.1\linewidth}
>{\centering\arraybackslash}m{0.1\linewidth}
>{\centering\arraybackslash}m{0.1\linewidth}
>{\centering\arraybackslash}m{0.1\linewidth}
>{\centering\arraybackslash}m{0.105\linewidth}
>{\centering\arraybackslash}m{0.105\linewidth}
>{\centering\arraybackslash}m{0.11\linewidth}
>{\centering\arraybackslash}m{0.1\linewidth}
>{\centering\arraybackslash}m{0.1\linewidth}}
\multicolumn{9}{c}{\includegraphics[width=0.99\linewidth]{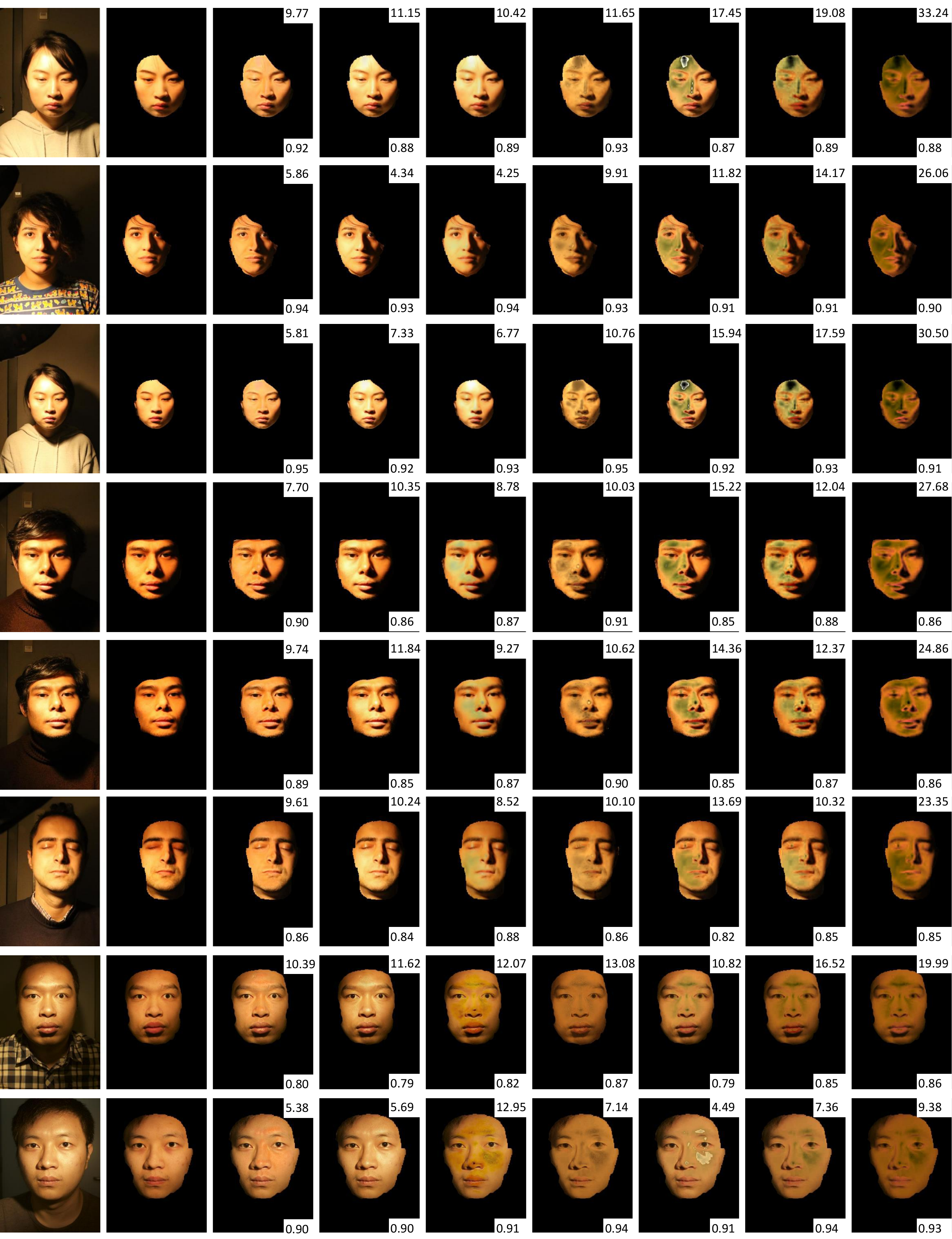}}\\
(a) & (b) & (c) & (d) & (e) & (f) & (g) & (h)&(i)\\
\end{tabular}
\caption{Additional highlight removal comparisons on laboratory images with ground truth. Face regions are cropped out automatically by landmark detection \cite{zhu2012face}. (a) Input photo. (b) Ground truth captured by cross-polarization for lab data. (c-h) Highlight removal results by (c) our finetuned Highlight-Net, (d) Highlight-Net without finetuning, (e) \cite{shi2017learning}, (f) \cite{li2017specular}, (g) \cite{shen2013real}, (h) \cite{yang2010real}, and (i) \cite{tan2004separating}. RMSE values are given at the top-right, and SSIM at the bottom-right. RMSE and SSIM are computed on highlight layers. }
\label{fig:highlightcompare2}
\end{figure}

\begin{figure}
\centering
\begin{tabular}
{>{\centering\arraybackslash}m{0.1\linewidth}
>{\centering\arraybackslash}m{0.1\linewidth}
>{\centering\arraybackslash}m{0.1\linewidth}
>{\centering\arraybackslash}m{0.1\linewidth}
>{\centering\arraybackslash}m{0.1\linewidth}
>{\centering\arraybackslash}m{0.1\linewidth}
>{\centering\arraybackslash}m{0.1\linewidth}
>{\centering\arraybackslash}m{0.1\linewidth}
>{\centering\arraybackslash}m{0.1\linewidth}}
\multicolumn{9}{c}{\includegraphics[width=0.99\linewidth]{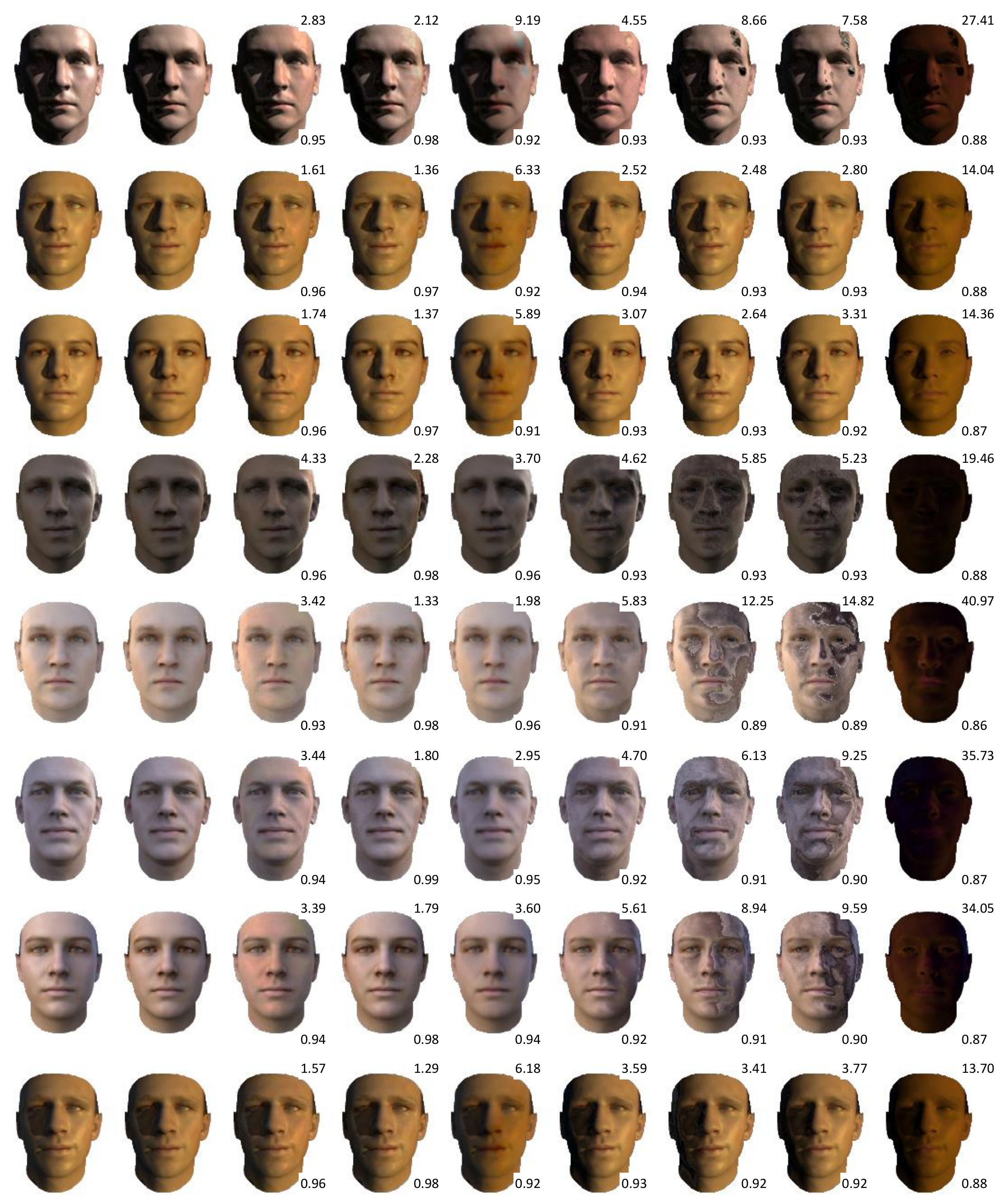}}\\
(a) & (b) & (c) & (d) & (e) & (f) & (g) &(h)&(i)\\
\end{tabular}
\caption{Highlight removal comparisons on a subset of the synthetic images. (a) Input photo. (b) Diffuse rendering under the same illumination. (c-h) Highlight removal results by (c) our method, (d) our pretrained net, (e) \cite{shi2017learning}, (f) \cite{li2017specular}, (g) \cite{shen2013real}, (h) \cite{yang2010real}, and (i) \cite{tan2004separating}. RMSE values are given at the top-right, and SSIM at the bottom-right. RMSE and SSIM are computed on highlight layers. }
\label{fig:highlightsynthetic}
\end{figure}

We also provide additional comparisons of highlight removal on laboratory images with ground truth, shown in Figure~\ref{fig:highlightcompare1} and Figure~\ref{fig:highlightcompare2}, with RMSE and SSIM values given in the figures. Our method mostly outperforms the previous techniques, which generally have difficulty in dealing with the saturated pixels that commonly appear in highlight regions. 
Comparisons on a subset of the synthetic data used in the quantitative evaluation are shown in Figure~\ref{fig:highlightsynthetic}. It can be seen that our method generates results similar to the diffuse renderings. The error histograms for quantitative evaluation on 100 synthetic faces and 30 real faces are shown in Table 1 in the main paper and Figure~\ref{fig:highlighthis}.

\begin{figure}
\centering
{\includegraphics[width=0.98\linewidth]{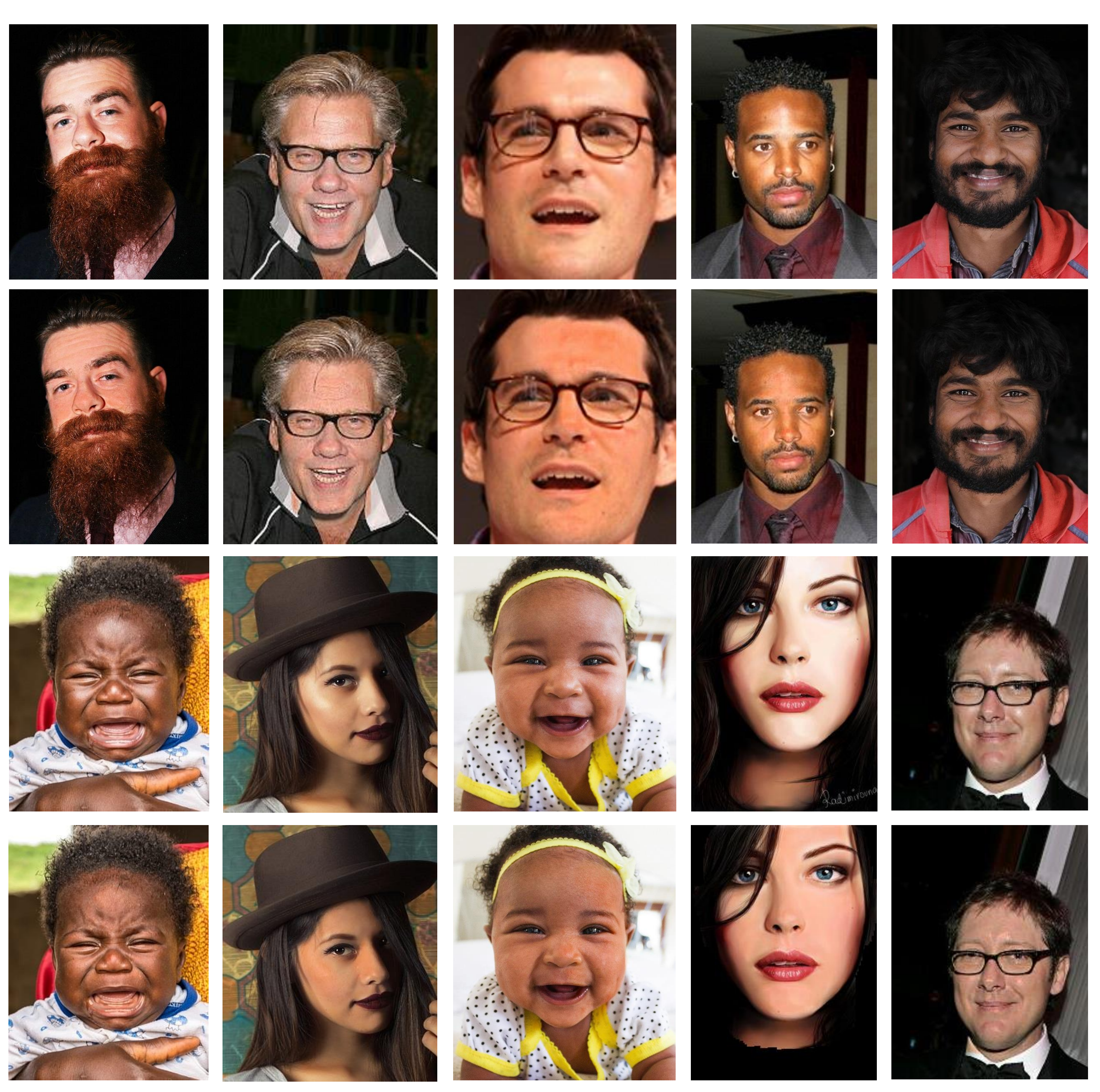}}
   \caption{Evaluation of highlight removal on testing data with non-neutral expressions, occluders and various ages/skin tones. Input images are shown on the first and third rows, corresponding highlight removal results are shown on the second and fourth rows. }
\label{fig:harddata}
\end{figure}

To show the robustness of Highlight-Net, we tested hard examples like non-neutral expressions, with occluders like glasses or beard, and various ages or skin tones, we provide additional results in Figure~\ref{fig:harddata}, which indicate reasonable performance. 
\section{Additional results on illumination estimation and virtual object insertion}

%\section{Illumination estimation images}

\begin{figure}
\centering
\includegraphics[width=\linewidth]{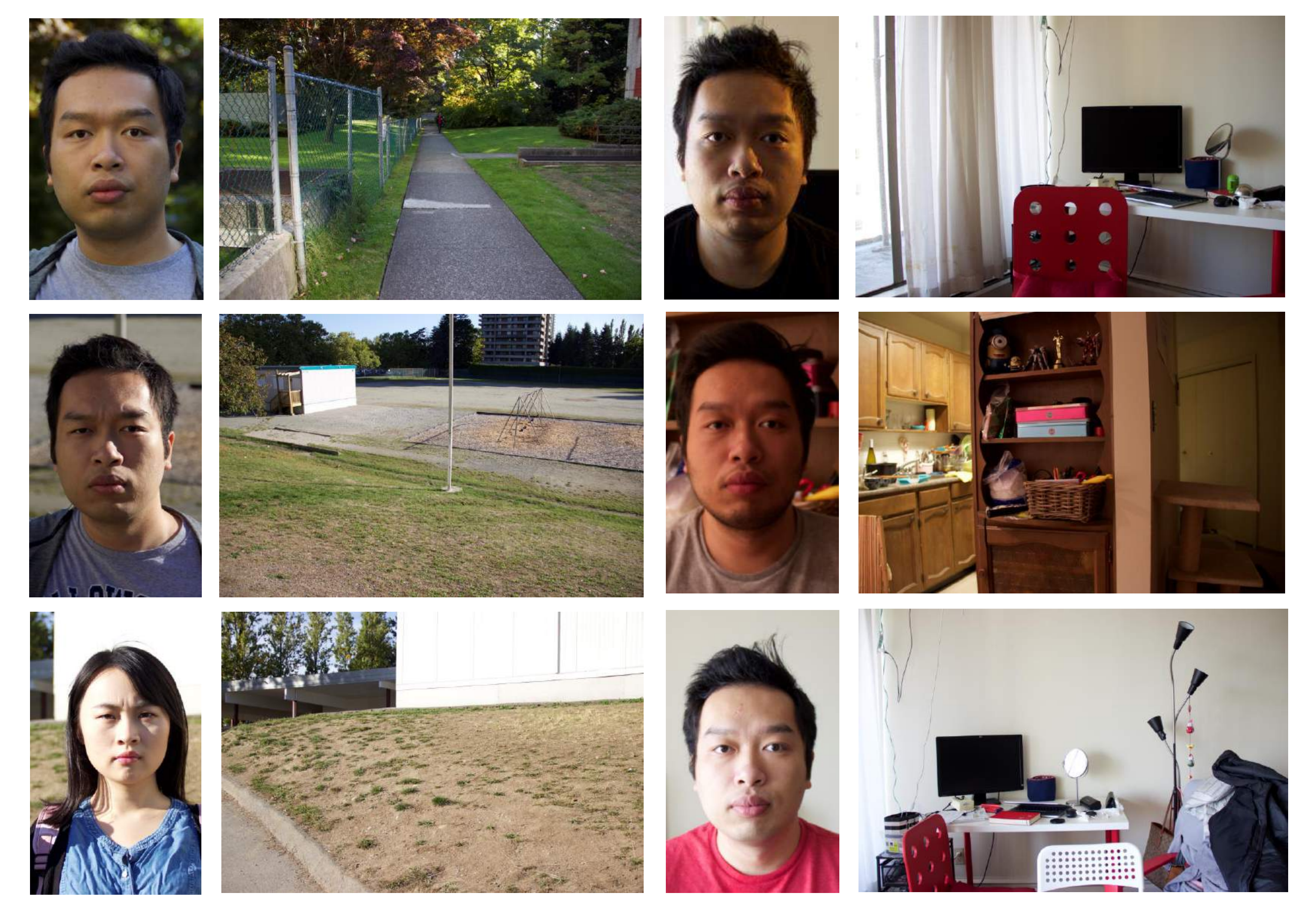}\
   \caption{Images used as input to \cite{gardner2017learning} (indoor scenes) and \cite{hold2016deep} (outdoor scenes). For each example, the left is the face image used as input for the other illumination estimation algorithms, and the right is the corresponding background photos used as input for \cite{gardner2017learning} and \cite{hold2016deep}.}
\label{fig:background}
\end{figure}

As mentioned in the main text, the methods in \cite{gardner2017learning,hold2016deep} are trained on images that do not contain people and that have a broader view of the scene. So in the experiments, we provide these methods with different input images that fit these characteristics, as shown in Figure \ref{fig:background}. Specifically, the person is removed, and a wider angle of the scene is captured. 

Also described in the main text, for our quantitative evaluation on illumination estimation with synthetic data for our method, \cite{lombardi2016reflectance} and \cite{knorr2014real}, we provided synthetic faces rendered under the ground truth environment maps as input images. We used 50 indoor and 50 outdoor environment maps in the quantitative evaluation, and 5 synthetic faces under each environment map as input images. In total, 500 synthetic faces are tested for the evaluation. For \cite{hold2016deep} and \cite{gardner2017learning}, we provided LDR photos cropped from the center of the ground truth environment maps as input images. For each of them, 50 outdoor/indoor LDR crops are tested for the evaluation. 

\begin{figure*}[t]
\centering
\begin{tabular}
{>{\centering\arraybackslash}m{0.2\linewidth}
>{\centering\arraybackslash}m{0.16\linewidth}
>{\centering\arraybackslash}m{0.16\linewidth}
>{\centering\arraybackslash}m{0.16\linewidth}
>{\centering\arraybackslash}m{0.17\linewidth}}
\multicolumn{5}{c}{\includegraphics[width=0.9\linewidth]{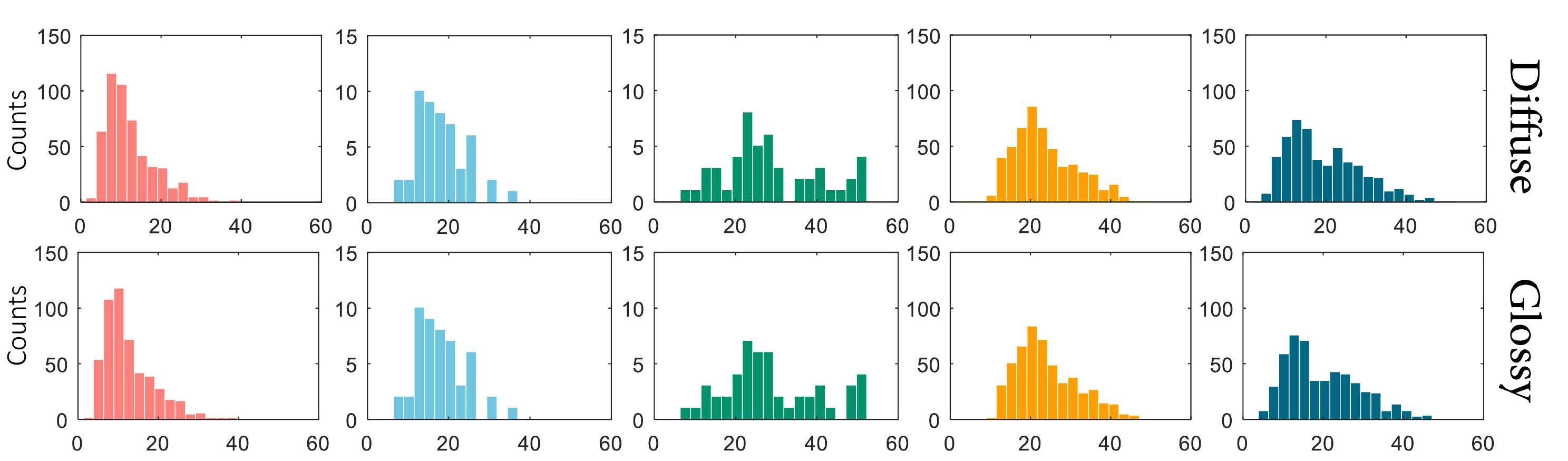}}\\
(a) & (b) & (c) & (d)&(e)\\
\end{tabular}
   \caption{Relighting RMSE histograms of a diffuse/glossy Stanford bunny lit by illumination estimated by (a) our method, (b) \cite{hold2016deep} (for outdoor scenes), (c) \cite{gardner2017learning} (for indoor scenes), (d) \cite{lombardi2016reflectance} and (e) \cite{knorr2014real} (spherical harmonics representation). Visual comparisons of the relighted diffuse/glossy bunnies are available in the supplement. }
\label{fig:bunny}
\end{figure*}

The evaluation is done by rendering a diffuse and a glossy Stanford bunny under ground truth and estimated environment maps, and computing the RMSE between these renderings. We use Keyshot \cite{keyshot} as the rendering engine, and for the diffuse bunny we set the diffuse reflectance as white and the specular reflectance as black (all zeros). For the glossy bunny, we choose ``hard shiny white plastic'' as the material and set the specular reflectance as white, the roughness factor to 0.004, and the refraction index to 1.362. Due to different scaling factors between HDR environment maps estimated by different methods, each diffuse rendering is normalized by its maximum value before computing the error, and the corresponding scaling factors of each environment map are also used for the glossy renderings. The relighting errors are shown in Table 2 in the main paper and Figure~\ref{fig:bunny}.

\begin{figure}
\centering
\begin{tabular}
{>{\centering\arraybackslash}m{0.14\linewidth}
>{\centering\arraybackslash}m{0.14\linewidth}
>{\centering\arraybackslash}m{0.17\linewidth}
>{\centering\arraybackslash}m{0.17\linewidth}
>{\centering\arraybackslash}m{0.17\linewidth}
>{\centering\arraybackslash}m{0.17\linewidth}}
\multicolumn{6}{c}{\includegraphics[width=0.99\linewidth]{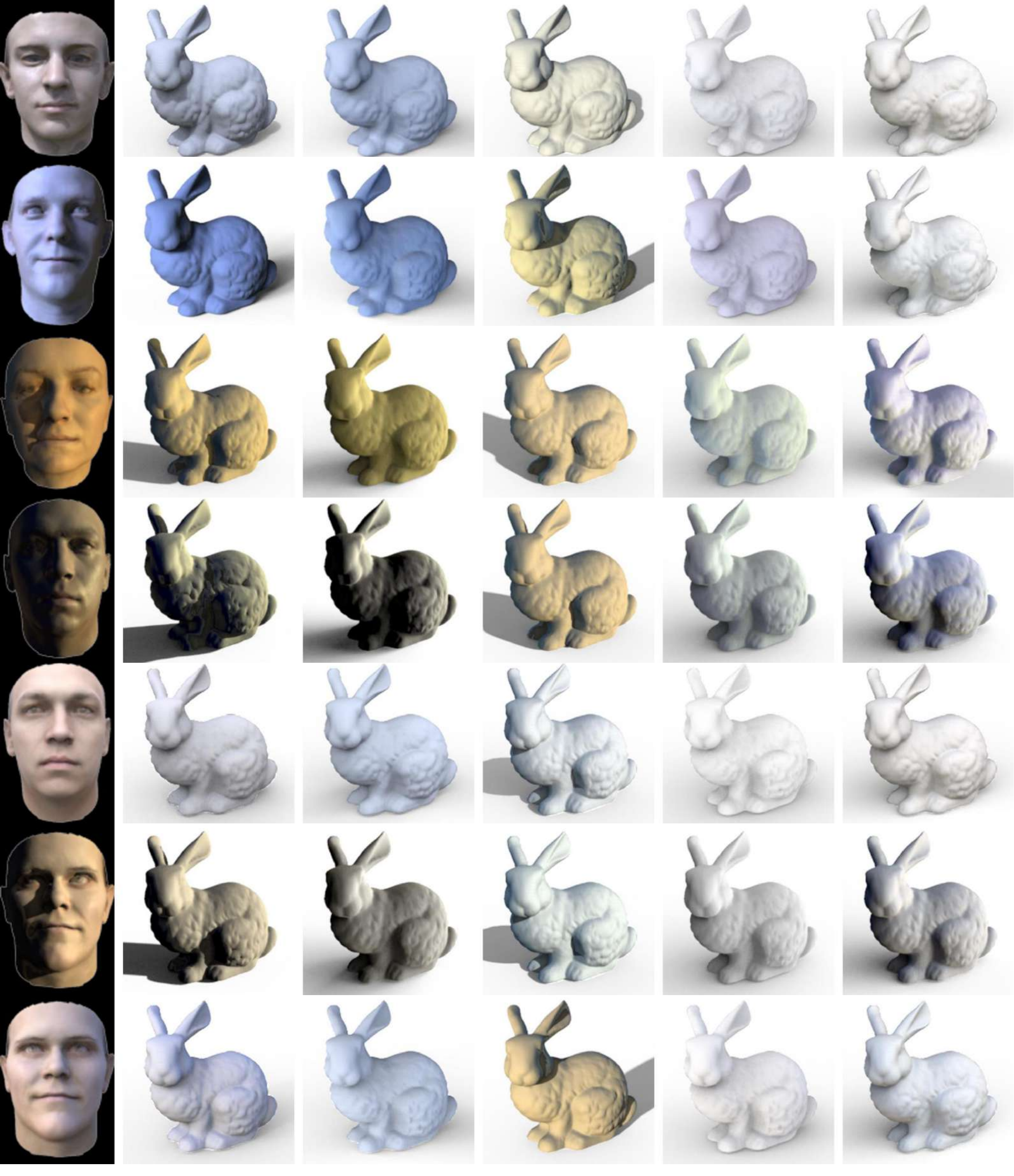}}\\
(a) & (b) & (c) & (d) & (e) & (f) \\
\end{tabular}
\caption{Comparisons of diffuse Stanford bunny relit by estimated outdoor illuminations. (a) Input photo. (b) Bunnies under ground truth environment maps. (c-f) Bunnies relit by environment maps estimated by (c) our method, (d) \cite{hold2016deep}, (e)\cite{lombardi2016reflectance} and (f) \cite{knorr2014real}. }
\label{fig:diffuseoutdoor1}
\end{figure}

\begin{figure}
\centering
\begin{tabular}
{>{\centering\arraybackslash}m{0.14\linewidth}
>{\centering\arraybackslash}m{0.14\linewidth}
>{\centering\arraybackslash}m{0.17\linewidth}
>{\centering\arraybackslash}m{0.17\linewidth}
>{\centering\arraybackslash}m{0.17\linewidth}
>{\centering\arraybackslash}m{0.17\linewidth}}
\multicolumn{6}{c}{\includegraphics[width=0.99\linewidth]{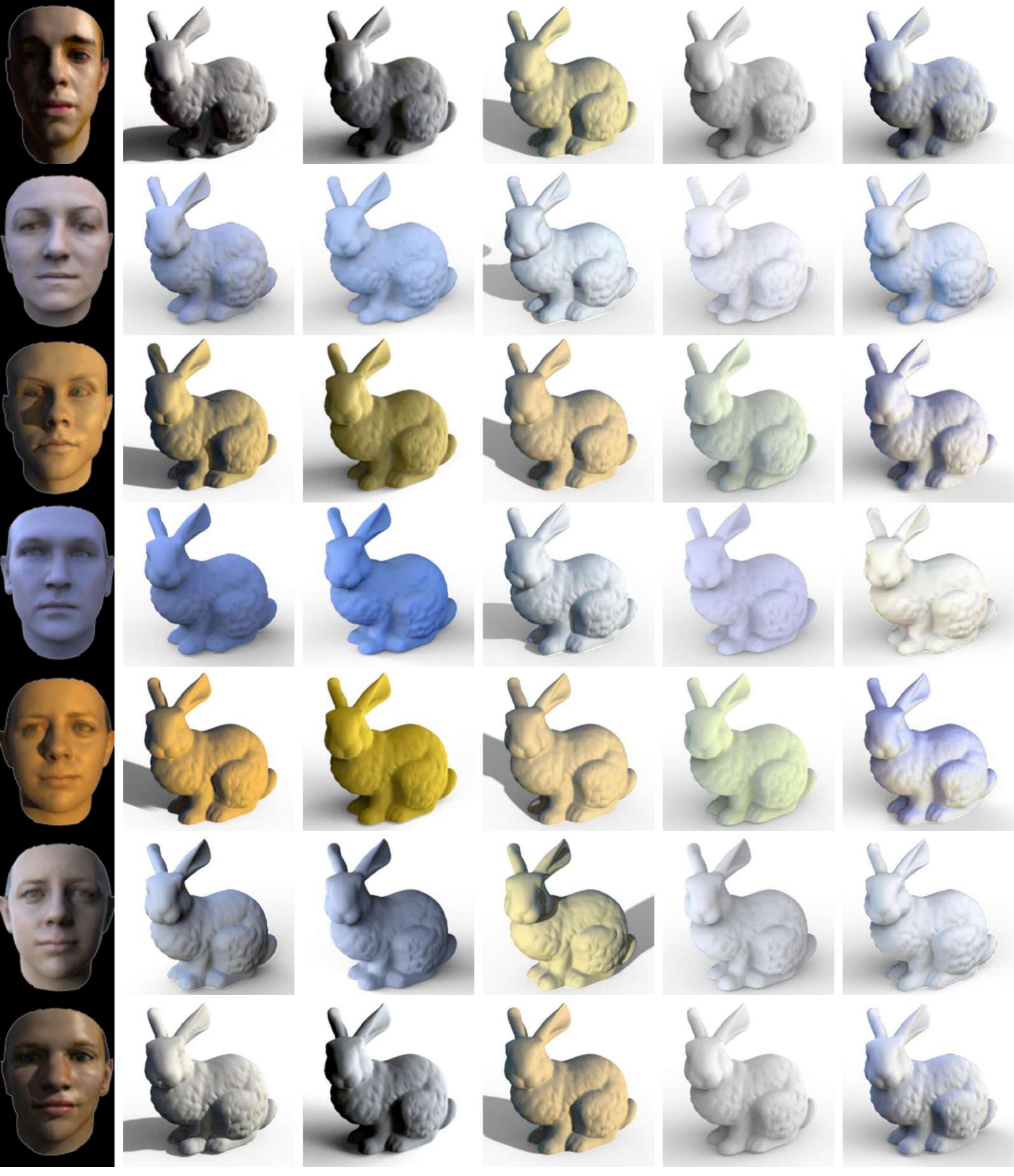}}\\
(a) & (b) & (c) & (d) & (e) & (f) \\
\end{tabular}
\caption{Comparisons of diffuse Stanford bunny relit by estimated outdoor illuminations. (a) Input photo. (b) Bunnies under ground truth environment maps. (c-f) Bunnies relit by environment maps estimated by (c) our method, (d) \cite{hold2016deep}, (e)\cite{lombardi2016reflectance} and (f) \cite{knorr2014real}.  }
\label{fig:diffuseoutdoor2}
\end{figure}

\begin{figure}
\centering
\begin{tabular}
{>{\centering\arraybackslash}m{0.14\linewidth}
>{\centering\arraybackslash}m{0.14\linewidth}
>{\centering\arraybackslash}m{0.17\linewidth}
>{\centering\arraybackslash}m{0.17\linewidth}
>{\centering\arraybackslash}m{0.17\linewidth}
>{\centering\arraybackslash}m{0.17\linewidth}}
\multicolumn{6}{c}{\includegraphics[width=0.99\linewidth]{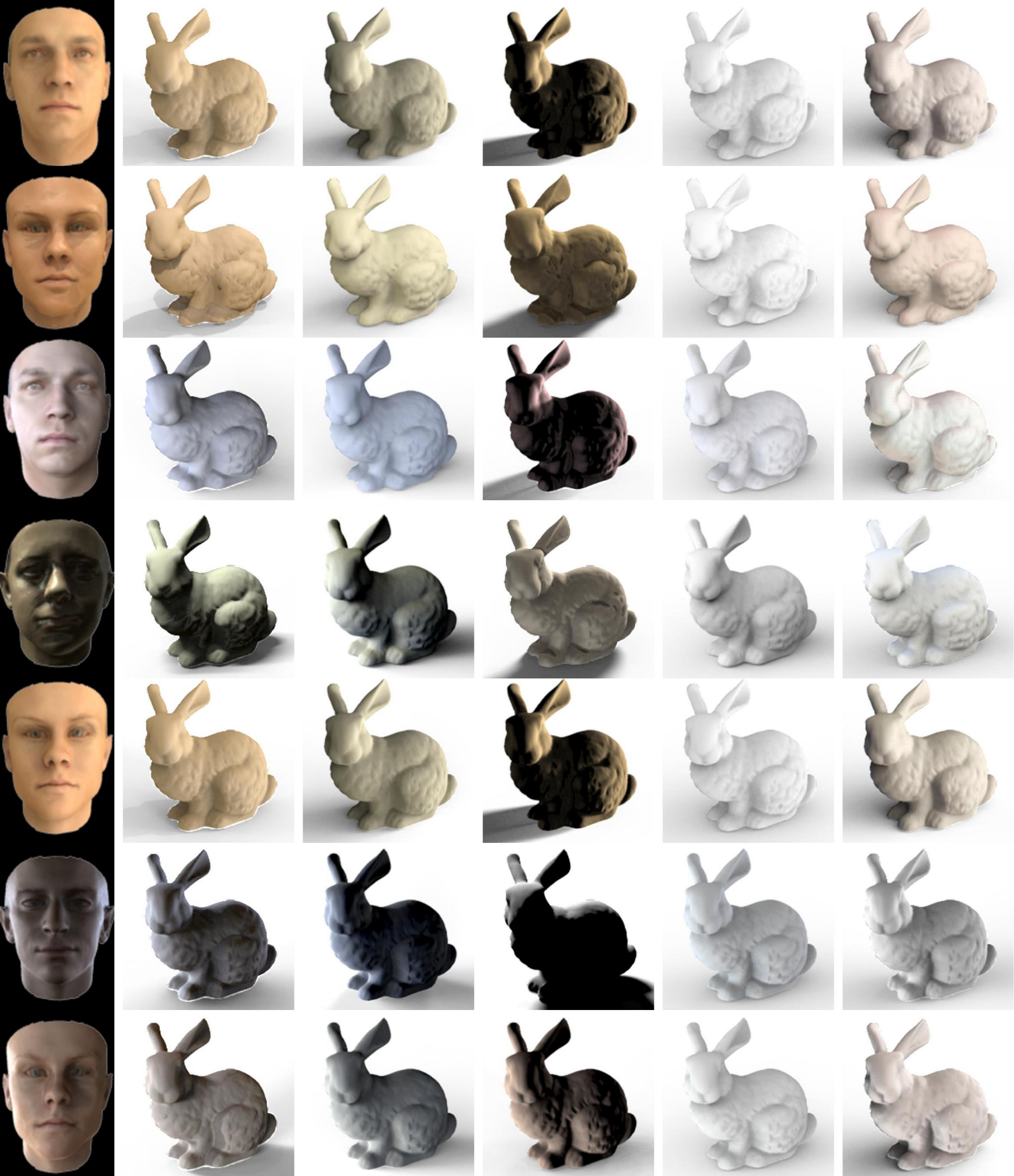}}\\
(a) & (b) & (c) & (d) & (e) & (f) \\
\end{tabular}
\caption{Comparisons of diffuse Stanford bunny relit by estimated indoor illuminations. (a) Input photo. (b) Bunnies under ground truth environment maps. (c-f) Bunnies relit by environment maps estimated by (c) our method, (d) \cite{gardner2017learning}, (e)\cite{lombardi2016reflectance} and (f) \cite{knorr2014real}.  }
\label{fig:diffuseindoor1}
\end{figure}

\begin{figure}
\centering
\begin{tabular}
{>{\centering\arraybackslash}m{0.14\linewidth}
>{\centering\arraybackslash}m{0.14\linewidth}
>{\centering\arraybackslash}m{0.17\linewidth}
>{\centering\arraybackslash}m{0.17\linewidth}
>{\centering\arraybackslash}m{0.17\linewidth}
>{\centering\arraybackslash}m{0.17\linewidth}}
\multicolumn{6}{c}{\includegraphics[width=0.99\linewidth]{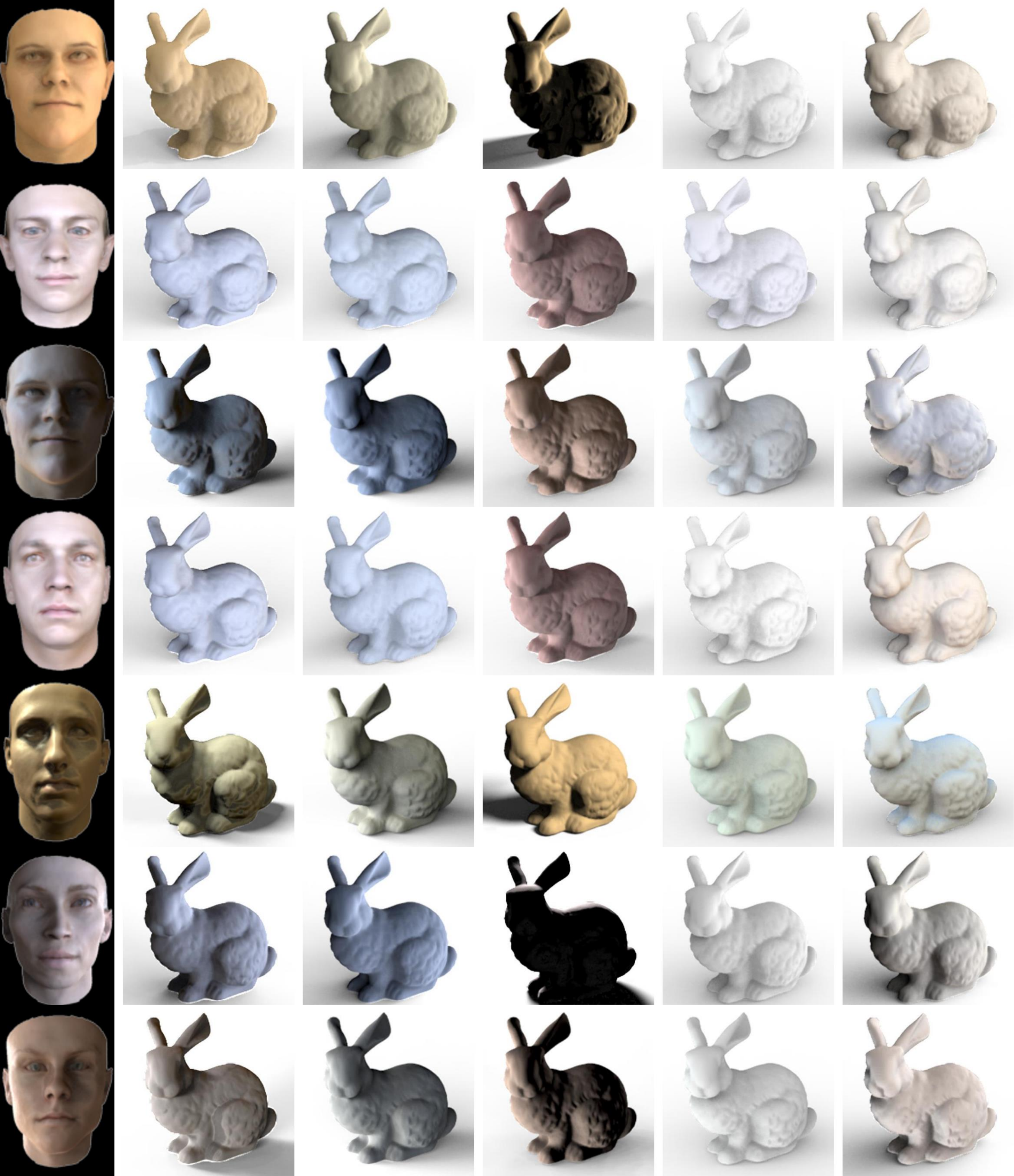}}\\
(a) & (b) & (c) & (d) & (e) & (f) \\
\end{tabular}
\caption{Comparisons of diffuse Stanford bunny relit by estimated indoor illuminations. (a) Input photo. (b) Bunnies under ground truth environment maps. (c-f) Bunnies relit by environment maps estimated by (c) our method, (d) \cite{gardner2017learning}, (e)\cite{lombardi2016reflectance} and (f) \cite{knorr2014real}.  }
\label{fig:diffuseindoor2}
\end{figure}

\begin{figure}
\centering
\begin{tabular}
{>{\centering\arraybackslash}m{0.14\linewidth}
>{\centering\arraybackslash}m{0.14\linewidth}
>{\centering\arraybackslash}m{0.17\linewidth}
>{\centering\arraybackslash}m{0.17\linewidth}
>{\centering\arraybackslash}m{0.17\linewidth}
>{\centering\arraybackslash}m{0.17\linewidth}}
\multicolumn{6}{c}{\includegraphics[width=0.99\linewidth]{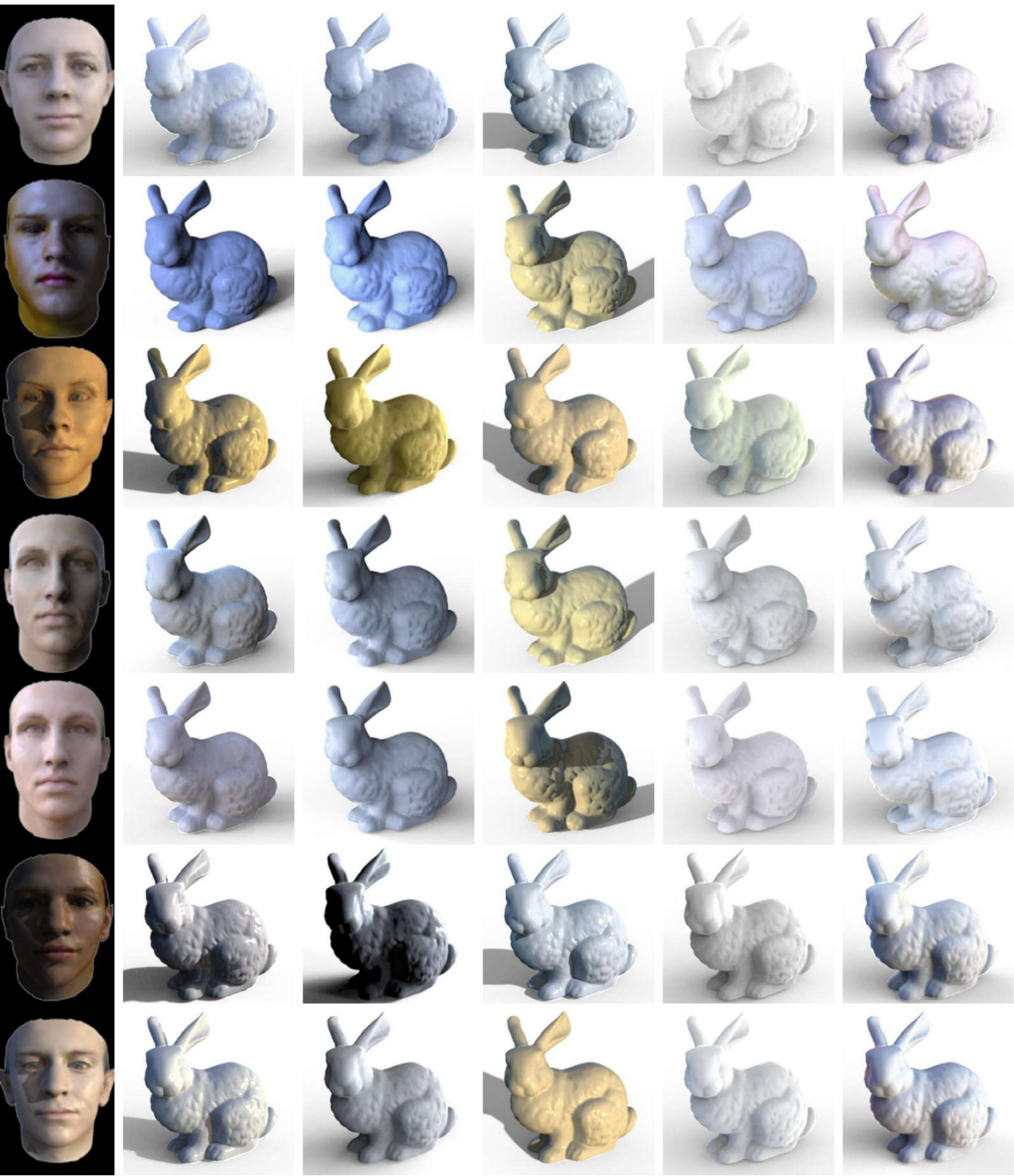}}\\
(a) & (b) & (c) & (d) & (e) & (f) \\
\end{tabular}
\caption{Comparisons of glossy Stanford bunny relit by estimated outdoor illuminations. (a) Input photo. (b) Bunnies under ground truth environment maps. (c-f) Bunnies relit by environment maps estimated by (c) our method, (d) \cite{hold2016deep}, (e)\cite{lombardi2016reflectance} and (f) \cite{knorr2014real}.  }
\label{fig:glossyoutdoor1}
\end{figure}

\begin{figure}
\centering
\begin{tabular}
{>{\centering\arraybackslash}m{0.14\linewidth}
>{\centering\arraybackslash}m{0.14\linewidth}
>{\centering\arraybackslash}m{0.17\linewidth}
>{\centering\arraybackslash}m{0.17\linewidth}
>{\centering\arraybackslash}m{0.17\linewidth}
>{\centering\arraybackslash}m{0.17\linewidth}}
\multicolumn{6}{c}{\includegraphics[width=0.99\linewidth]{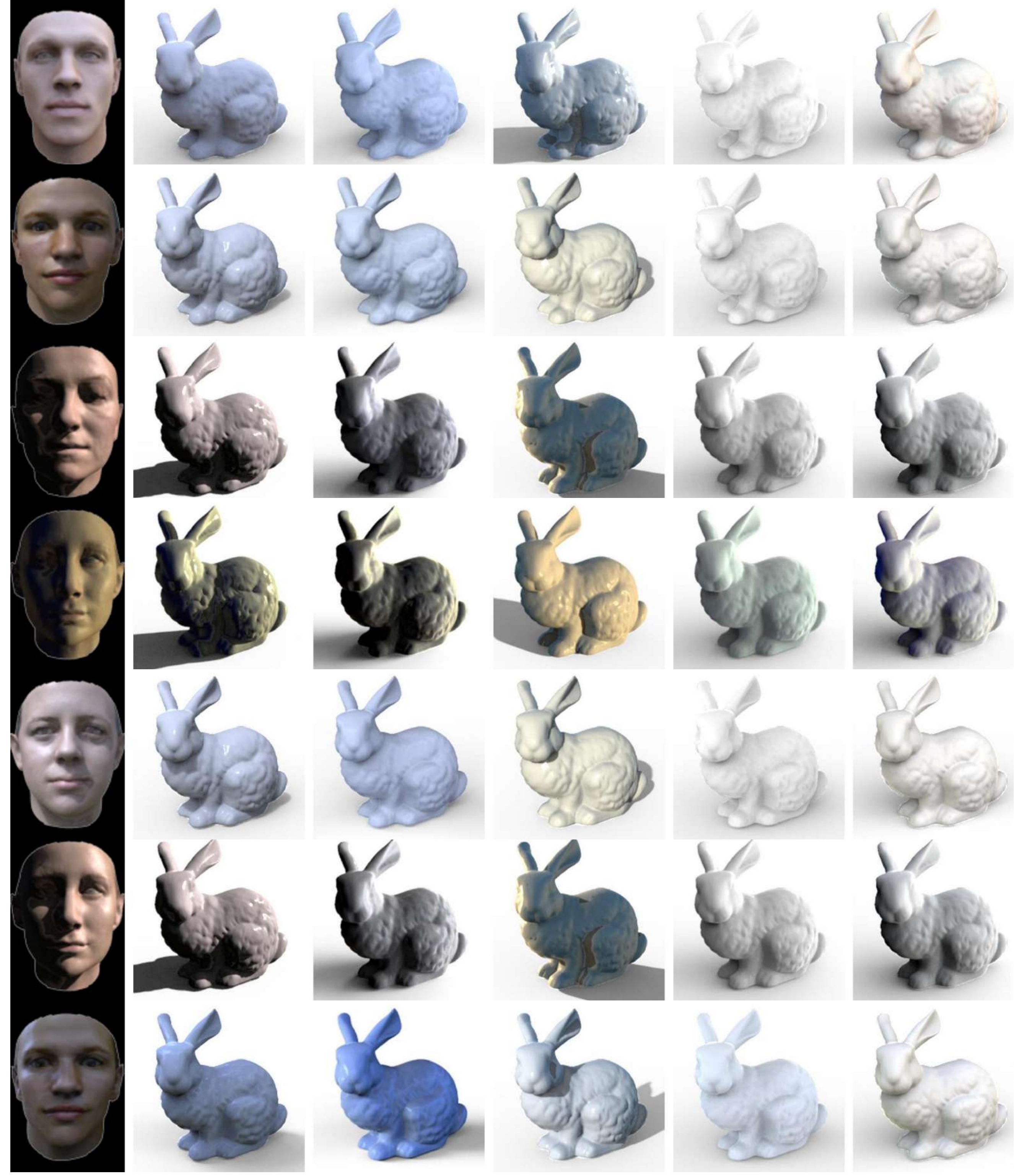}}\\
(a) & (b) & (c) & (d) & (e) & (f) \\
\end{tabular}
\caption{Comparisons of glossy Stanford bunny relit by estimated outdoor illuminations. (a) Input photo. (b) Bunnies under ground truth environment maps. (c-f) Bunnies relit by environment maps estimated by (c) our method, (d) \cite{hold2016deep}, (e)\cite{lombardi2016reflectance} and (f) \cite{knorr2014real}.  }
\label{fig:glossyoutdoor2}
\end{figure}

\begin{figure}
\centering
\begin{tabular}
{>{\centering\arraybackslash}m{0.14\linewidth}
>{\centering\arraybackslash}m{0.14\linewidth}
>{\centering\arraybackslash}m{0.17\linewidth}
>{\centering\arraybackslash}m{0.17\linewidth}
>{\centering\arraybackslash}m{0.17\linewidth}
>{\centering\arraybackslash}m{0.17\linewidth}}
\multicolumn{6}{c}{\includegraphics[width=0.99\linewidth]{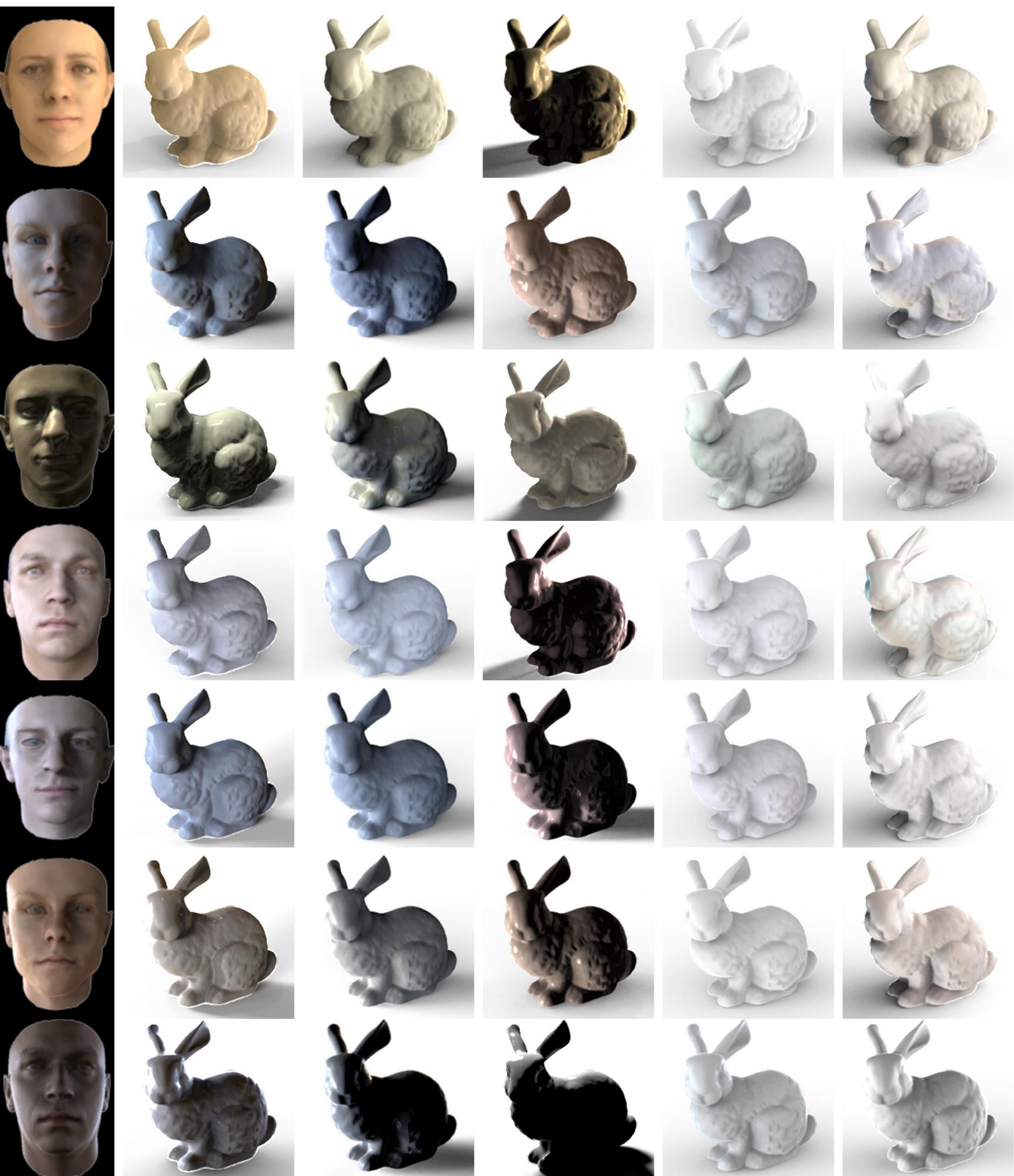}}\\
(a) & (b) & (c) & (d) & (e) & (f) \\
\end{tabular}
\caption{Comparisons of glossy Stanford bunny relit by estimated indoor illuminations. (a) Input photo. (b) Bunnies under ground truth environment maps. (c-f) Bunnies relit by environment maps estimated by (c) our method, (d) \cite{gardner2017learning}, (e)\cite{lombardi2016reflectance} and (f) \cite{knorr2014real}.  }
\label{fig:glossyindoor1}
\end{figure}

\begin{figure}
\centering
\begin{tabular}
{>{\centering\arraybackslash}m{0.14\linewidth}
>{\centering\arraybackslash}m{0.14\linewidth}
>{\centering\arraybackslash}m{0.17\linewidth}
>{\centering\arraybackslash}m{0.17\linewidth}
>{\centering\arraybackslash}m{0.17\linewidth}
>{\centering\arraybackslash}m{0.17\linewidth}}
\multicolumn{6}{c}{\includegraphics[width=0.99\linewidth]{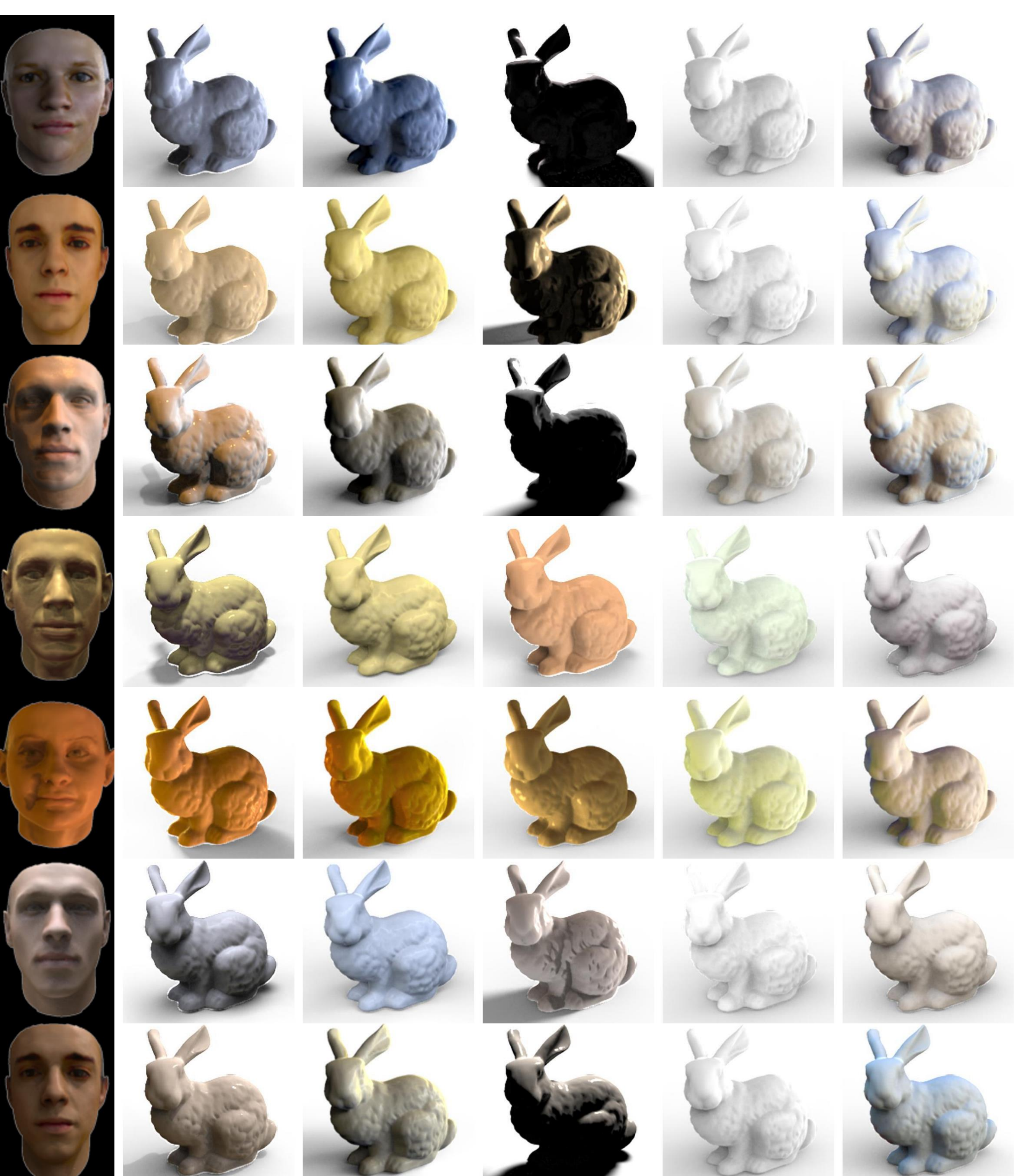}}\\
(a) & (b) & (c) & (d) & (e) & (f) \\
\end{tabular}
\caption{Comparisons of glossy Stanford bunny relit by estimated indoor illuminations. (a) Input photo. (b) Bunnies under ground truth environment maps. (c-f) Bunnies relit by environment maps estimated by (c) our method, (d) \cite{gardner2017learning}, (e)\cite{lombardi2016reflectance} and (f) \cite{knorr2014real}.  }
\label{fig:glossyindoor2}
\end{figure}

\begin{figure}
\centering
\begin{tabular}
{>{\centering\arraybackslash}m{0.18\linewidth}
>{\centering\arraybackslash}m{0.2\linewidth}
>{\centering\arraybackslash}m{0.2\linewidth}
>{\centering\arraybackslash}m{0.2\linewidth}
>{\centering\arraybackslash}m{0.18\linewidth}}
\multicolumn{5}{c}{\includegraphics[width=0.99\linewidth]{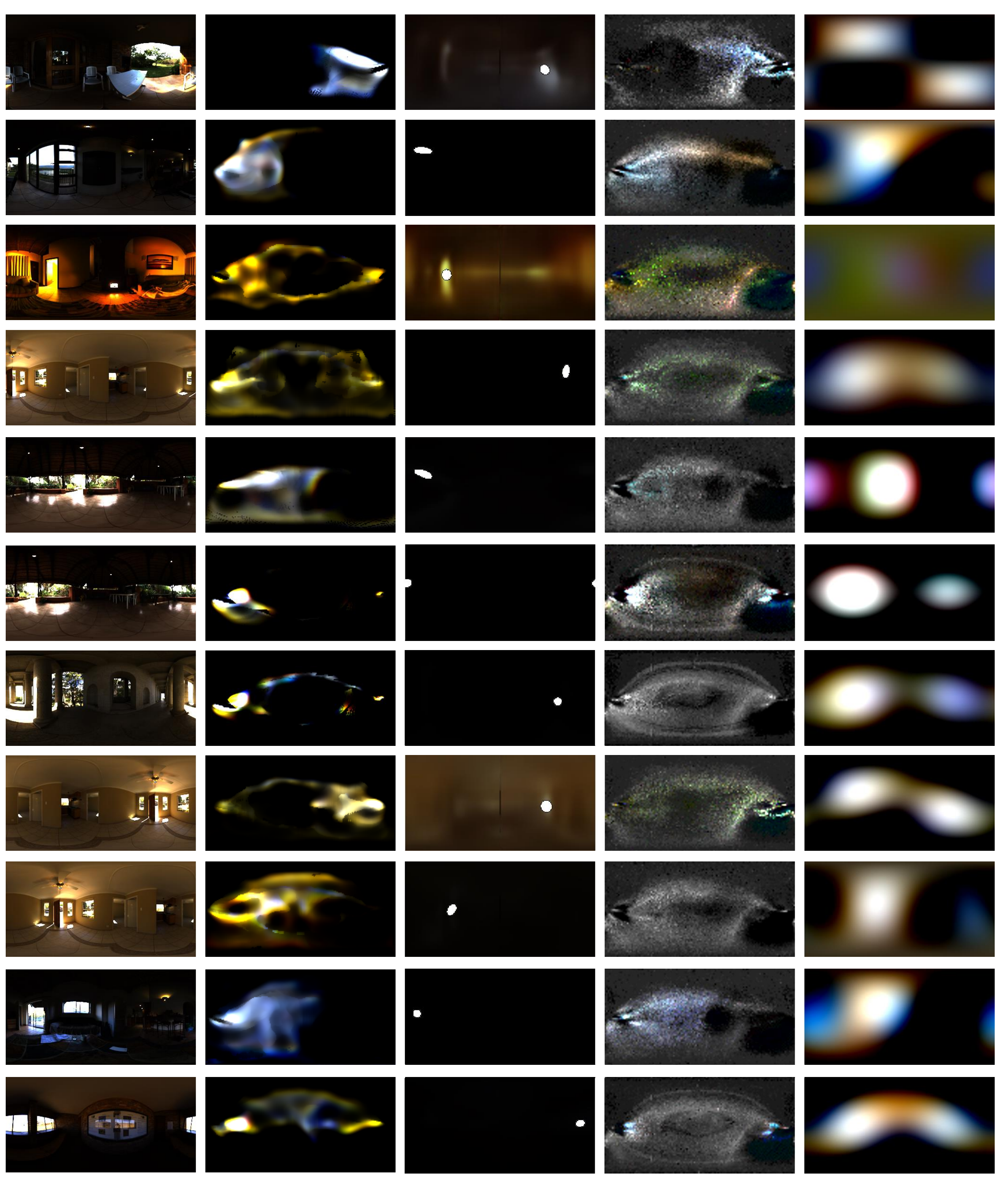}}\\
(a) & (b) & (c) & (d) & (e) \\
\end{tabular}
\caption{Comparisons of selected indoor data used in quantitative evaluation of illumination estimation. (a) Ground truth indoor environment maps, (b-e) indoor environment maps estimated by (b) our method, (c) \cite{gardner2017learning}, (d)\cite{lombardi2016reflectance} and (e) \cite{knorr2014real}. Total intensities of all environment maps are normalized to be the same. }
\label{fig:indoormaps}
\end{figure}

\begin{figure}
\centering
\begin{tabular}
{>{\centering\arraybackslash}m{0.18\linewidth}
>{\centering\arraybackslash}m{0.2\linewidth}
>{\centering\arraybackslash}m{0.2\linewidth}
>{\centering\arraybackslash}m{0.2\linewidth}
>{\centering\arraybackslash}m{0.18\linewidth}}
\multicolumn{5}{c}{\includegraphics[width=0.99\linewidth]{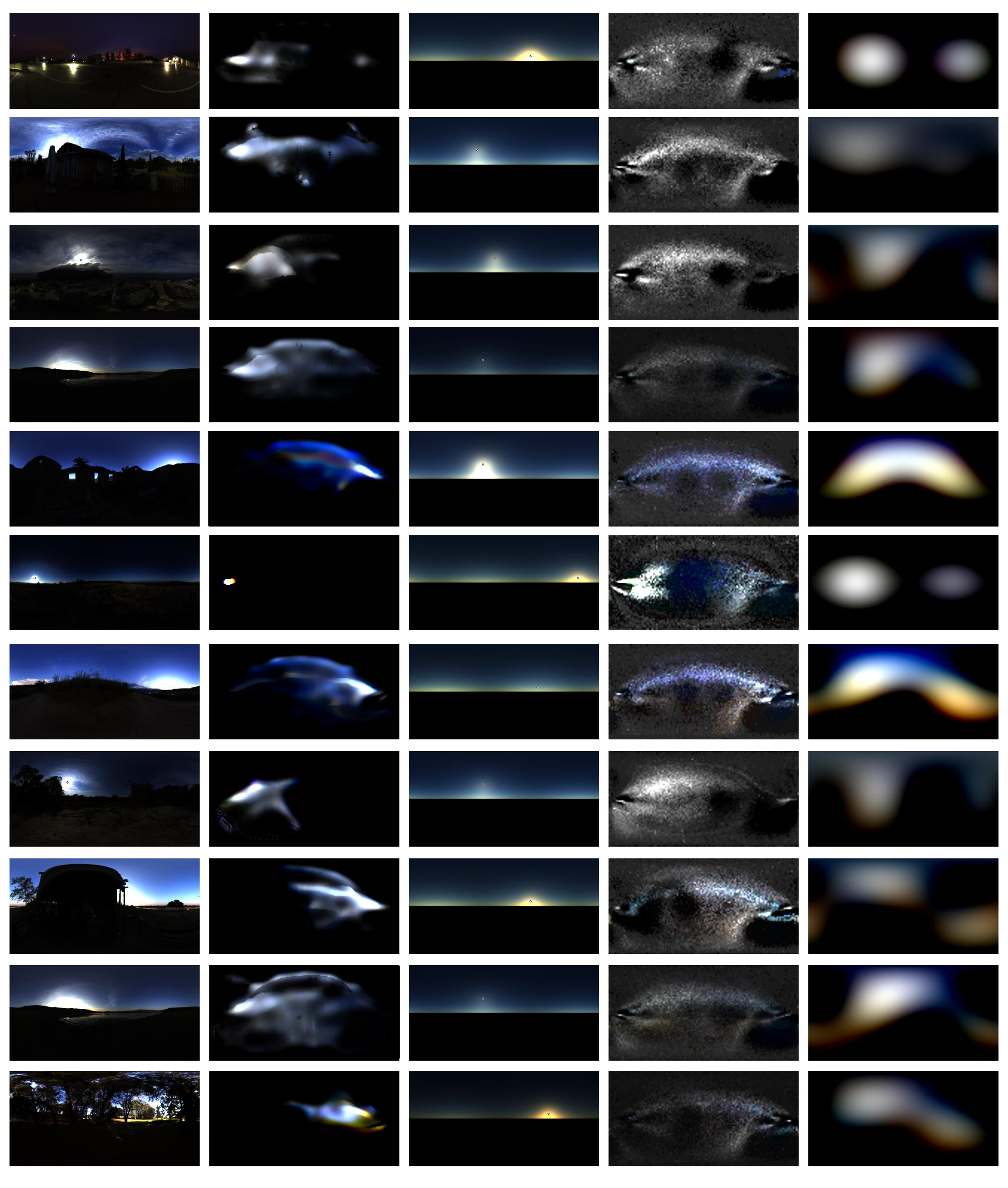}}\\
(a) & (b) & (c) & (d) & (e) \\
\end{tabular}
\caption{Comparisons of selected outdoor data used in quantitative evaluation of illumination estimation. (a) Ground truth indoor environment maps, (b-e) indoor environment maps estimated by (b) our method, (c) \cite{gardner2017learning}, (d)\cite{lombardi2016reflectance} and (e) \cite{knorr2014real}. Total intensities of all environment maps are normalized to be the same. }
\label{fig:outdoormaps}
\end{figure}

Comparisons on rendered diffuse bunnies under outdoor/indoor illuminations are shown in Figure \ref{fig:diffuseoutdoor1}-\ref{fig:diffuseoutdoor2} (outdoor), and Figure \ref{fig:diffuseindoor1}-\ref{fig:diffuseindoor2} (indoor). Comparisons on rendered glossy bunnies under outdoor/indoor illuminations are shown in Figure \ref{fig:glossyoutdoor1}-\ref{fig:glossyoutdoor2} (outdoor), and Figure \ref{fig:glossyindoor1}-\ref{fig:glossyindoor2} (indoor).
Comparisons on environment maps are displayed in Figure \ref{fig:indoormaps} for indoor scenes, and in Figure \ref{fig:outdoormaps} for outdoor scenes. 

\begin{figure}
\centering
{\includegraphics[width=0.98\linewidth]{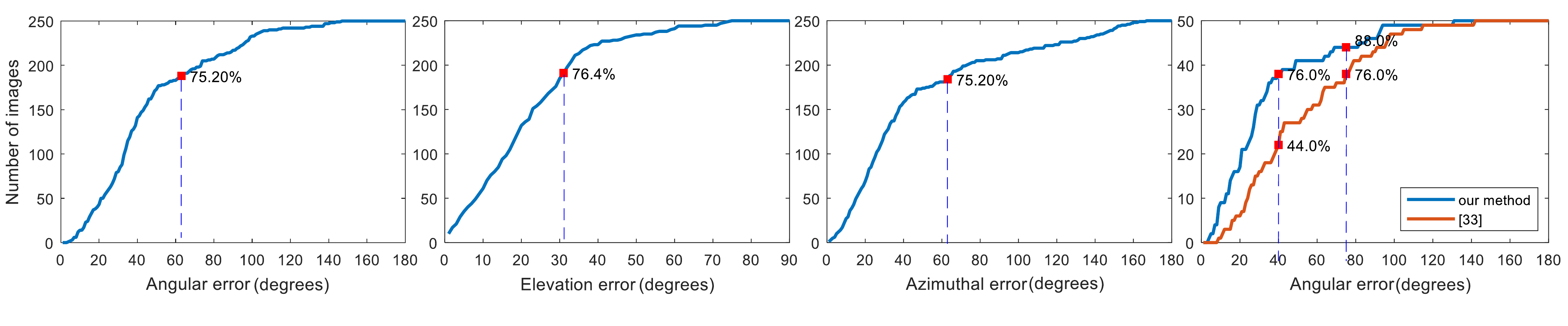}}
   \caption{Evaluation of sun position estimation on outdoor testing data. }
\label{fig:lightposition}
\end{figure}

To evaluate direction localization, we conducted an experiment on sun positions for outdoor scenes in Figure~\ref{fig:lightposition}, we computed the centroid of the predicted environment maps as the sun position, in terms of cumulative distribution of images w.r.t. error level as done in \cite{hold2016deep}, where the marked points indicate the error levels over more than $75\%$ of the testing data.

\begin{table}
\centering 
\begin{tabular}{c|c c c c c c c| c c c c c c} 
\hline 
Normalized RMSE & Ours & \cite{hold2016deep} & \cite{gardner2017learning} & \cite{lombardi2016reflectance} & \cite{knorr2014real}&\\ [0.5ex] 
\hline 
Mean (outdoor) &\bf{0.143}&0.163&$\backslash$&0.154&0.245\\ 
Mean (indoor) &\bf{0.045}&$\backslash$&0.050&0.083&0.286\\ 
\hline 
\end{tabular}
\caption{Errors in estimating environment maps from real data.} 
\label{table:real} 
\end{table}

%\begin{table}
%\centering 
%\begin{tabular}{c|c c c c c c c| c c c c c c} 
%\hline 
%Normalized RMSE & Ours & \cite{hold2016deep} & \cite{gardner2017learning} & \cite{lombardi2016reflectance} & \cite{knorr2014real}&\\ [0.5ex] 
%\hline 
%Mean &\bf{0.058}&0.126&0.086&0.092&0.282\\ 
%Median &\bf{0.042}&0.124&\bf{0.042}&0.082&0.275\\ 
%\hline 
%\end{tabular}
%\caption{Errors in estimating environment maps from real data.} 
%\label{table:real} 
%\end{table}

\begin{figure}
\centering
{\includegraphics[width=0.99\linewidth]{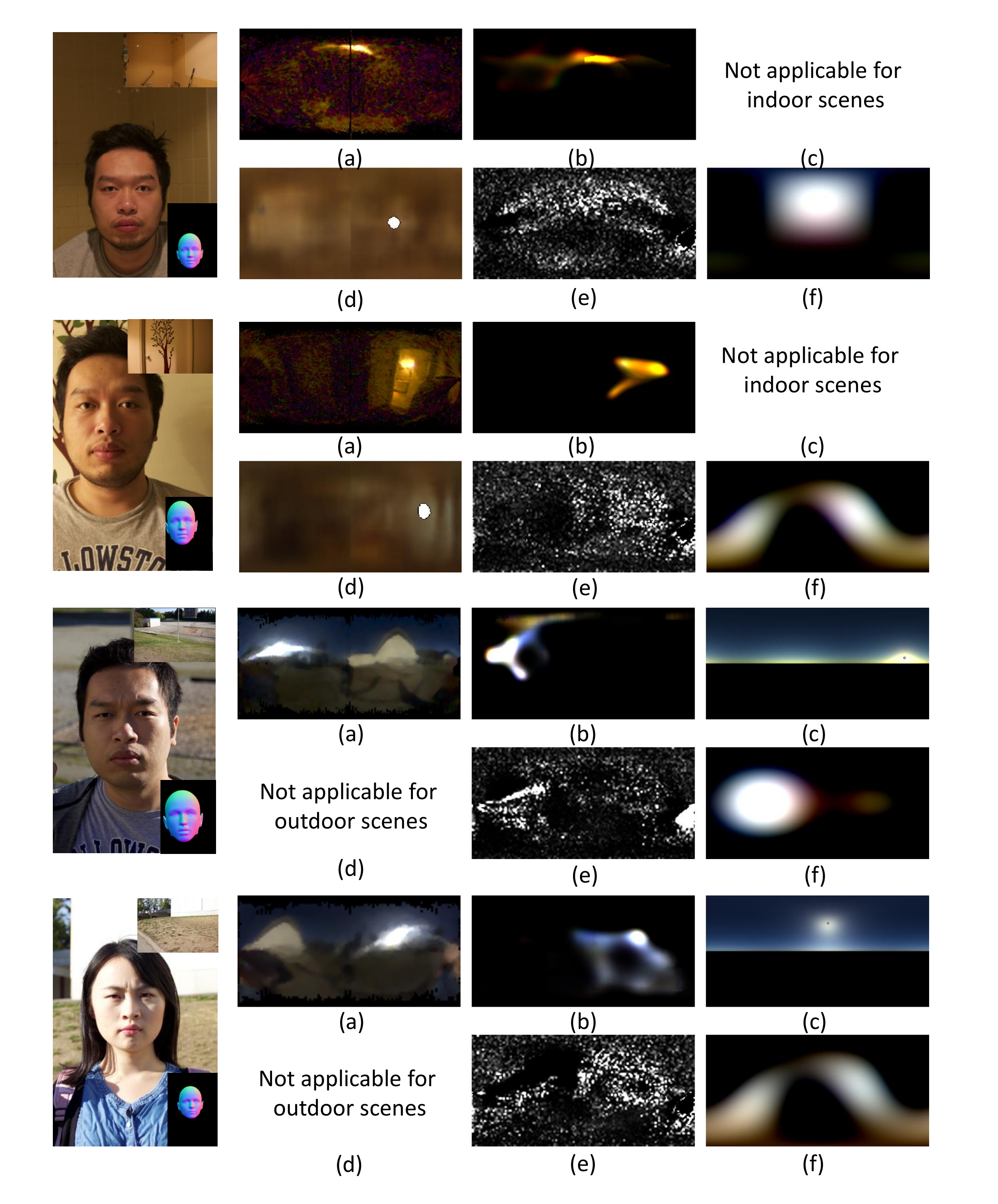}}\\
\caption{Comparisons of illumination estimation on real data. The faces on the left are input face photos. (a) Ground truth indoor environment maps, (b-f) indoor environment maps estimated by (b) our method, (c) \cite{hold2016deep}, (d)\cite{gardner2017learning}, (e)\cite{lombardi2016reflectance} and (f) \cite{knorr2014real}. Input background photos for \cite{gardner2017learning} and \cite{hold2016deep} are shown at top right of the input photos, and the input face normals for our method, \cite{lombardi2016reflectance} and \cite{knorr2014real} are shown at bottom right. }
\label{fig:realmaps1}
\end{figure}

\begin{figure}
\centering
{\includegraphics[width=0.99\linewidth]{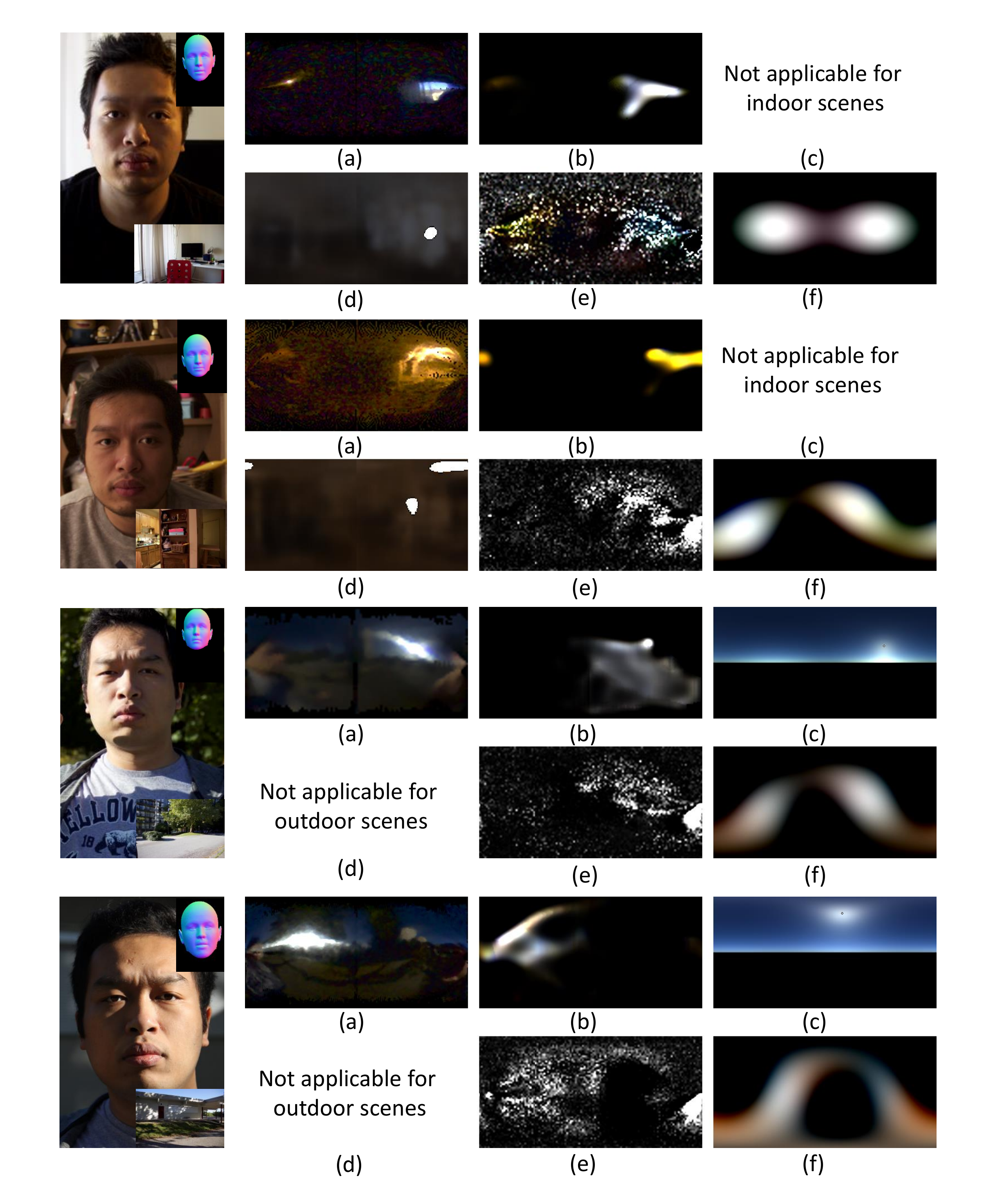}}\\
\caption{Comparisons of illumination estimation on real data. The faces on the left are input face photos. (a) Ground truth indoor environment maps, (b-f) indoor environment maps estimated by (b) our method, (c) \cite{hold2016deep}, (d)\cite{gardner2017learning}, (e)\cite{lombardi2016reflectance} and (f) \cite{knorr2014real}. Input background photos for \cite{gardner2017learning} and \cite{hold2016deep} are shown at bottom right of the input photos, and the input face normals for our method, \cite{lombardi2016reflectance} and \cite{knorr2014real} are shown at top right. }
\label{fig:realmaps2}
\end{figure}

For comparisons on real data, face images and their HDR environment maps are captured for 15 real scenes (7 indoor and 8 outdoor), with background images having faces excluded and a larger field of view for \cite{hold2016deep} and \cite{gardner2017learning}. Errors with respect to the captured ground truths are presented in Table \ref{table:real} in terms of RMSE normalized by the difference of the maximum and minimum intensity of the estimated environment map, which is commonly used to facilitate comparison between data with different scales, such as those from the intensity scaling factor of environment maps estimated by different methods. 
%Different from the synthetic data, the outdoor scenes in the real data usually have sunlight from the front of the face, such that the lighting source is out-of-view in the background images. In these cases, \cite{hold2016deep} is less precise for estimating sun positions. 
Visual comparisons on estimated environment maps are shown in Figure \ref{fig:realmaps1} and \ref{fig:realmaps2}. The methods in \cite{hold2016deep} and \cite{gardner2017learning} are applicable only to outdoor and indoor scenes, respectively. They were found to be generally less precise in estimating light source directions when light sources are out-of-view in background images, though they provide reasonable approximations. As seen in (e), the method in \cite{lombardi2016reflectance} may be relatively sensitive to imprecise geometry and surface textures. In (f), estimates of a low-order SH model are seen to lack detail. Our results in (b) most closely match the ground truth, with some error due partly to inexact estimation of face geometry. 

Additional comparisons on virtual object insertion are presented in Figure \ref{fig:outdoor}, where an outdoor scene is at the top and an indoor scene is at the bottom.

All codes will be publicly available shortly.

\begin{figure}
\centering
\begin{tabular}
{>{\centering\arraybackslash}m{0.2\linewidth}
>{\centering\arraybackslash}m{0.2\linewidth}
>{\centering\arraybackslash}m{0.18\linewidth}
>{\centering\arraybackslash}m{0.2\linewidth}
>{\centering\arraybackslash}m{0.2\linewidth}}
\multicolumn{5}{c}{\includegraphics[width=\linewidth]{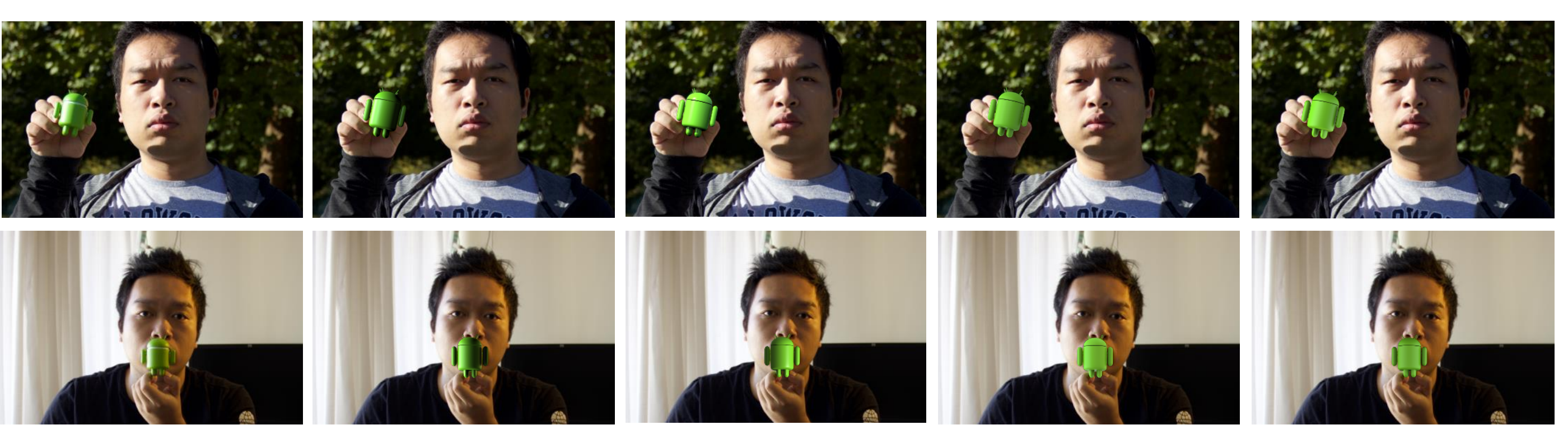}}\\
(a) & (b) & (c)&(d)&(e) \\
\end{tabular}
   \caption{Comparisons of object insertion results for outdoor (top) and indoor (bottom) scenes. (a) Photos containing the real object; (b) our method; (c) outdoor result by \cite{hold2016deep} and indoor result by \cite{gardner2017learning}; (d) \cite{lombardi2016reflectance}; (e) \cite{knorr2014real}. }
\label{fig:outdoor}
\end{figure}




\end{document}